\documentclass[10pt,journal,compsoc]{IEEEtran}
\ifCLASSOPTIONcompsoc
  \usepackage[nocompress]{cite}
\else
  \usepackage{cite}
\fi
\ifCLASSINFOpdf
\else
\fi
\usepackage{graphicx}
\usepackage{multirow}
\usepackage{makecell}
\usepackage{amsmath}
\usepackage{wrapfig}
\usepackage{booktabs}
\usepackage{titletoc}
\usepackage{tabularx}
\usepackage{color}
\usepackage{amsfonts}
\usepackage{amssymb} 
\usepackage{hyperref}

\ifCLASSOPTIONcompsoc
 \usepackage[caption=false,font=footnotesize,labelfont=sf,textfont=sf]{subfig}
\else
 \usepackage[caption=false,font=footnotesize]{subfig}
\fi

\begin{document}
%
\title{NIERT: Accurate Numerical Interpolation through Unifying Scattered Data Representations using Transformer Encoder}
%
%
%
%

\author{Shizhe~Ding,
        Boyang~Xia,
        Milong~Ren,
        Dongbo~Bu
\IEEEcompsocitemizethanks{\IEEEcompsocthanksitem Shizhe Ding, Boyang Xia, Milong Ren are with Key Lab of Intelligent Information Processing of Chinese Academy of Sciences (CAS), Institute of Computing Technology, CAS, Beijing 100190, China, and University of Chinese Academy of Sciences, Beijing 100049, China. \protect\\
E-mail: \{dingshizhe19s, xiaboyang20s, renmilong21b\}@ict.ac.cn
\IEEEcompsocthanksitem Dongbo Bu is with Key Lab of Intelligent Information Processing of Chinese Academy of Sciences (CAS), Institute of Computing Technology, CAS, Beijing 100190, China, University of Chinese Academy of Sciences, Beijing 100049, China, and Zhongke Big Data Academy, Zhengzhou 450046, China. \protect\\
E-mail: {dbu}@ict.ac.cn
\IEEEcompsocthanksitem Corresponding Author: Dongbo Bu
}
\thanks{Manuscript received March 8, 2022.}}

%
%

\markboth{Journal of \LaTeX\ Class Files,~Vol.~14, No.~8, August~2015}%
{Shell \MakeLowercase{\textit{et al.}}: Bare Demo of IEEEtran.cls for Computer Society Journals}
%



\IEEEtitleabstractindextext{%

\begin{abstract}

Interpolation for scattered data is a classical problem in numerical analysis, with a long history of theoretical and practical contributions. Recent advances have utilized deep neural networks to construct interpolators, exhibiting excellent and generalizable performance. However, they still fall short in two aspects: \textbf{1) inadequate representation learning}, resulting from separate embeddings of observed and target points in popular encoder-decoder frameworks and \textbf{2) limited generalization power}, caused by overlooking prior interpolation knowledge shared across different domains. To overcome these limitations, we present a \textbf{N}umerical \textbf{I}nterpolation approach using \textbf{E}ncoder \textbf{R}epresentation of \textbf{T}ransformers (called \textbf{NIERT}). On one hand, NIERT utilizes an encoder-only framework rather than the encoder-decoder structure. This way, NIERT can embed observed and target points into a unified encoder representation space, thus effectively exploiting the correlations among them and obtaining more precise representations.  On the other hand, we propose to pre-train NIERT on large-scale synthetic mathematical functions to acquire prior interpolation knowledge, and transfer it to multiple interpolation domains with consistent performance gain. On both synthetic and real-world datasets, NIERT outperforms the existing approaches by a large margin, i.e., 4.3$\sim$14.3$\times$ lower MAE on TFRD subsets, and 1.7/1.8/8.7$\times$ lower MSE on Mathit/PhysioNet/PTV datasets. The source code of NIERT is available at \url{https://github.com/DingShizhe/NIERT}.

\end{abstract}

}

\maketitle

\IEEEdisplaynontitleabstractindextext

%
\IEEEpeerreviewmaketitle

\IEEEraisesectionheading{\section{Introduction}\label{sec:introduction}}

%
%
%
%

\begin{figure}[!t]
    \centering
    \includegraphics[width=9cm]{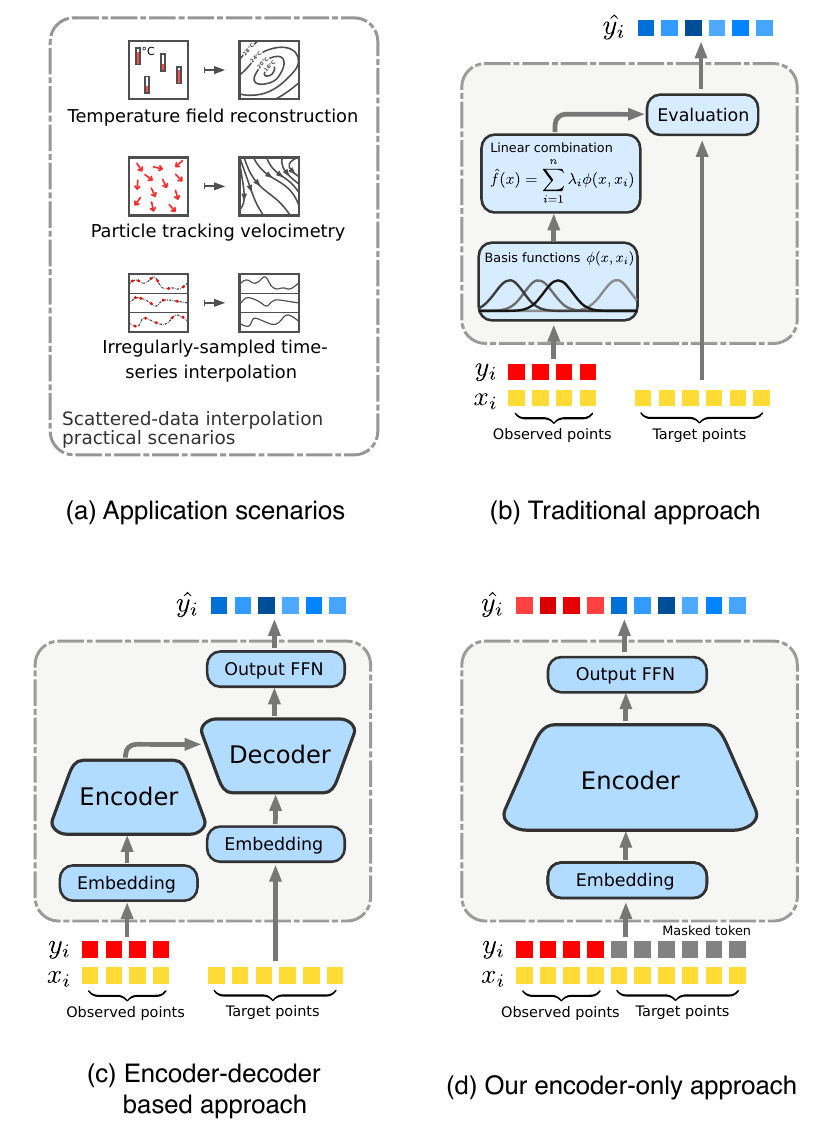}
    \caption{Application scenarios of scattered data interpolation, and comparison of existing approaches with our NIERT approach. (a) Various scenarios of scattered data interpolation. (b) The traditional approaches output explicit target functions expressed as a linear combination of basis functions. (c) In the encoder-decoder framework, the encoder and decoder process observed points and target points in a separate fashion, which results in inadequate representation learning. 
    (d) Our encoder-only framework processes observed and target points in a unified fashion, which helps obtain more precise target point representations.}
    \label{fig:background}%
\end{figure}

\IEEEPARstart{S}{cattered} data refers to a set of points and their corresponding values, in which the points lack structured relationships beyond their relative positions~\cite{franke1991scattered}.
Such data commonly arise in a wide range of theoretical and practical scenarios, including solving partial differential equations (PDEs) \cite{franke1991scattered,liu2016overview}, temperature field reconstruction \cite{chen2021machine}, particle tracking velocimetry \cite{dabiri2020particle}, and irregularly-sampled time series analysis \cite{lepot2017interpolation, shukla2019interpolation} (Fig.~\ref{fig:background}a).
Numerical interpolation is often required for scattered data, which involves estimating values for target points based on the exact values at some observed points.
For instance, in temperature field reconstruction for micro-scale electronics, interpolation methods are employed to obtain real-time working environments for electronic components from limited measurements, and inaccurate interpolation can significantly increase the cost of predictive maintenance.
Therefore, accurate numerical interpolation approaches are highly desirable.

A large number of approaches have been proposed for interpolating scattered data.
Traditional approaches utilize a linear combination of pre-defined basis functions to approximate the target function, as demonstrated in Figure~\ref{fig:background}b \cite{heath2018scientific}.
To adapt to different scenarios, various types of basis functions have been devised. 
For example, the spline interpolation algorithm uses polynomials of a small degree over subintervals of the approximation domain as basis functions, and the RBF interpolation algorithm uses radial basis functions.
Most of these schemes can theoretically guarantee interpolation accuracy when sufficient observed points are available; however, they have also been shown to be ineffective for sparse data points or complex functions \cite{bulirsch2002introduction}. Furthermore, these methods are inherently not learning-based, \textit{i.e.,} they cannot take advantage of data from the same function distribution, which limits their generalizability and interpolation accuracy.

Recent advances have exhibited an alternative strategy that uses neural networks to learn target functions from the given observed points.
For example, conditional neural processes (CNPs) \cite{pmlr-v80-garnelo18a} and their extensions \cite{kim2018attentive, NEURIPS2020_492114f6, lee2020residual} use neural networks to model the conditional distribution of regression functions given the observed points, and \cite{chen2021machine} proposed to use Transformer \cite{vaswani2017attention} to solve interpolation tasks in temperature field reconstruction. 
Unlike traditional approaches that directly output explicit target functions, these neural models take the observed points and target points' locations as input, and estimate the values of these target points.
Thus, it is reasonable to view this task as a set-to-set task, which motivates all of these models to employ encoder-decoder structures (Fig.~\ref{fig:background}c).
The encoder here embeds the observed points to a latent space while the decoder estimates values for target points. 
Compared with the traditional approaches, these learning-based methods can learn the function distribution as well as the correlation among scattered points from existing datasets and generalize to new interpolation tasks, which significantly improves the interpolation accuracy.


Despite the excellent performance of existing learning-based interpolation models for scattered data, they still fall short in two aspects: 
\textbf{1) overlooking the homogeneity of observed and target points}, which leads to inadequate representation learning of scattered points, and \textbf{2) lack of prior interpolation knowledge}, which leads to limited generalization power.
The underlying reasons of these limitations are: First, the encoder-decoder-based models process the observed and target points in the encoder and decoder separately, which keeps their learned representations in different feature spaces. However, the observed points and target points are intrinsically homogeneous, \textit{i.e.}, they are all sampled from the same function. Treating observed and target points separately  may  hinder the transfer of information from observed points to target points and increase the difficulty of learning the correlations between the observed points and target points.
This strategy makes the representation of target points  not precise enough, which limits the final interpolation accuracy.

Second, another shortcoming of these learning-based interpolation models is the lack of prior interpolation knowledge. 
The continuity and smoothness of the target function are always considered in traditional scattered data interpolation algorithms, \textit{e.g.,} using smooth basis functions to obtain smooth target functions \cite{franke1991scattered}.
This prior knowledge reflects the general properties of the function distribution, effectively narrowing the potential range of the target function and guaranteeing interpolation accuracy \cite{heath2018scientific}.
However, these existing neural interpolators lack such prior knowledge since they learn interpolation knowledge only from a domain-specific scattered dataset. This makes them lack generalization power to various domains.

To address these limitations, we propose a highly-accurate learning-based numerical interpolation approach for scattered data, which can effectively learn precise point representations with high generalization power. The basic ideas of our approach is as follows: 
First, to obtain more precise representations of the target points, we propose a novel interpolation model, called \textbf{NIERT}, which uses a transformer-based encoder-only structure rather than  popular encode-decoder structures. Different from previous works, NIERT processes both observed and target points in a unified fashion (see Fig.~\ref{fig:background}d). To bridge the gap between target points and observed points, \textit{i.e.}, to address the missing values of target points, we augment target points with learnable mask tokens, which is similar to BERT \cite{devlin2018bert}. Then we feed them into the encoder together with the observed points. This design improves modeling the correlations between these two types of points, which yields more precise representations and improves interpolation interpretability\footnote{The higher ``interpretability'' here means more visually reasonable correlations between observed and target points (See Sec. \ref{subsec_contrib} for details.)}.
Moreover, to prevent the unexpected interference posed by target points on the observed points, we introduce a partial self-attention mechanism to calculate attention among the given data points except for the influence of target points on the observed points at each layer. This further boosts interpolation robustness.



Second, we use the pre-training technique to incorporate the  prior interpolation knowledge into our NIERT interpolator, which has been successfully used for representation learning \cite{devlin2018bert,bao2021beit}. The key step towards this goal is building a task set for scattered-data interpolation with rich prior interpolation knowledge. For this purpose, we synthesized a large number of symbolic mathematical functions and generated a new task set by sampling 
from these functions. Large-scale syntheses of mathematical functions have the potential to cover real-world continuous and smooth functions in many domains at low cost. 
With this enhancement, the pre-trained NIERT interpolators can be easily transferred to a variety of domain-specific tasks with a lower learning difficulty and higher interpolation accuracy.




The main contributions of this study are summarized as follows. 

\begin{itemize}
    \item We design a transformer-based encoder-only interpolation framework and embed the observed and target points into a unified feature space, to obtain more accurate target point representations. Also, we propose a partial self-attention mechanism with a strong inductive bias for interpolation tasks, to avoid the interference of one type of points onto the others, which also boosts robustness.
    
    

    
    \item We leverage the pre-training technique to improve the generalization power of interpolation models. We synthesize a large-scale interpolation task set by generating a massive amount of symbolic mathematical functions. {To the best of our knowledge}, this study is the first work to propose the pre-trained models for scatter-data interpolation. 
    \item  Our approach significantly outperforms both state-of-the-art learning-based approaches and traditional approaches on 4 synthetic and real-world datasets in terms of both interpolation accuracy and interpretability, which shows the potential of our approach in a wide range of application fields. 

    
\end{itemize}

The paper is structured as follows: Section~\ref{sec:related} provides a literature review. Section~\ref{sec:method} presents our approach. Section~\ref{sec:exp_setting} elaborates on the experimental setup. Section~\ref{sec:exp_setting} reports  experimental results and analysis. Section~\ref{sec:discussion} discusses the connections between our approach and traditional interpolation methods. Finally, Section~\ref{sec:conclusion} draws conclusions.

\begin{figure*}
    \centering
    \includegraphics[width=\textwidth]{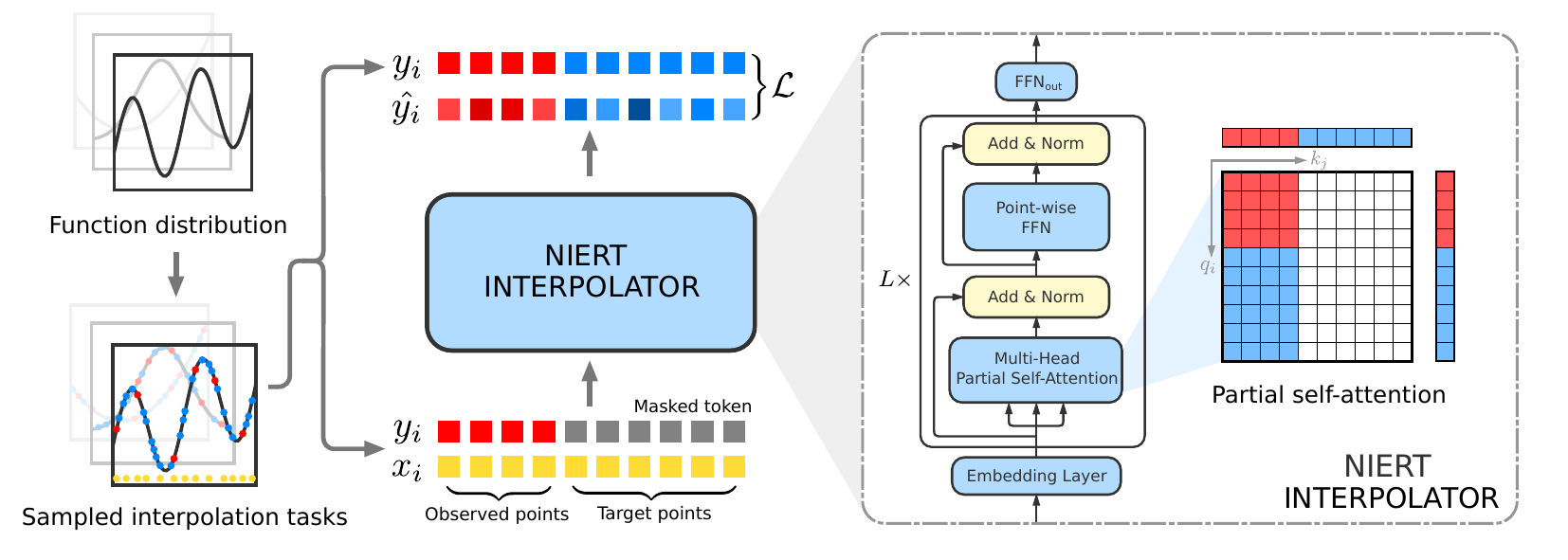}
    \caption{Overview of NIERT training process. Here, $x_i$ represents the position of a point, and  $y_i$ represents its value. The predicted values of the point is denoted as $\hat{y_i}$. We prepare the training interpolation tasks by first sampling functions from a function distribution $\mathcal{F}$ and then sampling observed points $O$ and target points $T$ on each function. NIERT trains an interpolator over this data. The partial self-attention mechanism facilitates exploiting the correlations between observed points and target points and avoiding the unexpected interference of target points on observed points.}
    \label{fig:pipeline}
\end{figure*}

\section{Related works}
\label{sec:related}

\subsection{Traditional interpolation approaches for  scattered data}

Traditional interpolation approaches for scattered data construct interpolation functions through combining  explicitly pre-defined  basis functions. The representative approaches include Lagrange interpolation, Newton interpolation \cite{heath2018scientific}, B-spline interpolation \cite{hall1976optimal}, Shepard’s method \cite{gordon1978shepard}, Kriging \cite{wackernagel2003multivariate}, and radial basis function interpolation (RBF) \cite{powell1987radial, fornberg2007runge}. Among these approaches, the classical  Lagrange interpolation, Newton interpolation and B-splines interpolation are usually used for univariate interpolation. \cite{wang2010high} proposed a high-order multivariate approximation scheme for scattered data sets, in which approximation error is represented using Taylor expansions at data points, and basis functions are determined by minimizing the approximation error.

\subsection{Neural network-based interpolation approaches}

Equipped with deep neural networks, data-driven interpolation and reconstruction methods show great advantages and potential.
For instance, convolutional neural networks (CNNs)  have been applied in the interpolation tasks of single image super-resolution \cite{tai2017image, li2021beginner}, and recurrent neural networks (RNNs) and Transformers have been used for interpolation of sequences like time series data \cite{shukla2019interpolation, shukla2021multitime}.

Recently, \cite{pmlr-v80-garnelo18a} proposed to model the conditional distribution of regression functions given observed points. The proposed approach, called conditional neural processes (CNPs), has shown increased estimation accuracy and generalizing ability. A variety of studies have been performed to enhance CNPs:  \cite{kim2018attentive} designed  attentive neural processes (ANPs), which shows improved accuracy. \cite{lee2020residual} leveraged Bayesian last layer (BLL) \cite{weber2018optimizing} for faster training and better prediction. In addition, the bootstrap technique was also employed  for further improvement \cite{NEURIPS2020_492114f6}.
To solve the interpolation task in 2D temperature field reconstruction, \cite{chen2021machine} proposed a Transformer-based approach, referred to as TFR-transformer, which can also be applied to solve interpolation tasks for scattered data with higher dimensions. 




\subsection{Masked language/image models and the pre-training technique}
The design of NIERT is also inspired by the recent advances in masked language/image models \cite{devlin2018bert,bao2021beit, NEURIPS2020_1457c0d6, he2021masked} and pre-trained models for symbolic regression \cite{biggio2021neural, valipour2021symbolicgpt}. The mask mechanism enables the models to reconstruct missing data from their context and effectively learn representations of language and images.
As a way of introducing prior knowledge, large-scale pre-training allows language or visual models to acquire more general text or image representation capabilities, which assists the learning process on downstream tasks and significantly improves their performance.
In addition, the pre-trained models for symbolic regression were designed to learn the map from scattered data to corresponding symbolic formulas~\cite{biggio2021neural,valipour2021symbolicgpt}.
Here, NIERT uses the mask mechanism to reconstruct the missing values of target points and uses the pre-training technique to acquire strong generalization power. 


\section{Method}
\label{sec:method}
\subsection{Overview of NIERT approach}

In the study, we focus on the interpolation task that can be formally described as follows: We are given $n$ observed points with known values $O = \{ (x_i, y_i) \}_{i=1}^{n}$, and $m$ target points with values to be determined, denoted as $T = \{ x_i \}_{i=n+1}^{n+m}$. Here, $x_i \in \mathbb{R}^{d_x}$ denotes the position of a point, $y_i=f(x_i) \in \mathbb{R}^{d_y}$ denotes the value of a point, and $f: \mathbb{R}^{d_x} \rightarrow \mathbb{R}^{d_y}$ denotes a function mapping positions to values. The $d_x$ and $d_y$ denote the dimension of a point's position and the dimension of the value, respectively.
The function $f$ is from a function distribution $\mathcal{F}$, which can be explicitly defined using a mathematical formula or implicitly represented using a set of scattered data in the form $(x_i, y_i)$. 
The goal of the interpolation task is to accurately estimate the values $f(x)$ for each target point $x\in T$ according to the observed points in $O$.

\begin{figure*}[!h]
    \centering
    \includegraphics[width=0.96\textwidth]{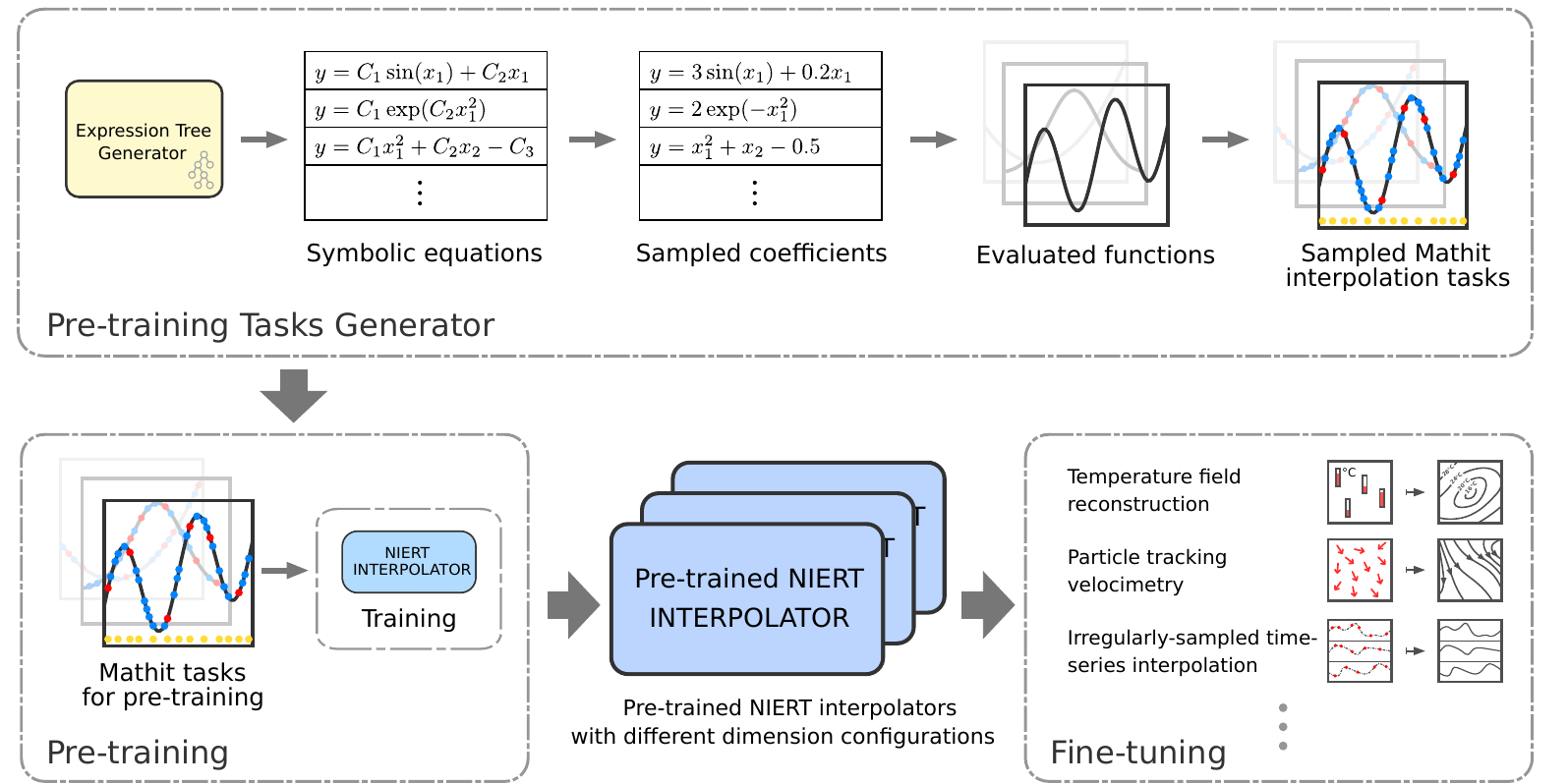}
    \caption{The generation process of \textbf{Mathit} pre-training task set and the pre-training and fine-tuning technique for NIERT interpolator. By randomly sampling the expression trees and coefficients, we synthesized a large number of mathematical functions that can represent a sufficiently wide and diverse function distribution. These functions are evaluated numerically and used to sample the Mathit interpolation tasks used for pre-training. The pre-training introduces prior knowledge into the interpolators and improves their generalization power in various real-world interpolation scenarios.}
    \label{fig:pretrain_pipeline}
\end{figure*}

The main element of our NIERT approach is a neural interpolator that learns to estimate values for target points. We train the neural interpolator using a set of interpolation tasks  sampled from the function distribution $\mathcal{F}$ (Sec. \ref{sec:arch_of_niert} and Fig.~\ref{fig:pipeline}). For each of the sampled interpolation tasks, we mask the values of target points, feed the task to the neural interpolator, and train  the interpolator to predict the values of target points and observed points as well. 
The training objective is to minimize the error between the predicted values and the ground-truth values of both target and observed points. The trained interpolator can be used to interpolate values of target points for the interpolation tasks from the distribution $\mathcal{F}$. In addition, to enable NIERT with generalization power to various interpolation domains,  we pre-train NIERT on a large-scale interpolation task set \textbf{Mathit}, which we generate by sampling on synthesized mathematical symbolic functions (see Sec. \ref{sec:pretrain} and Fig.~\ref{fig:pretrain_pipeline}). In the following sections, we first present the NIERT encoder-only structure, then we introduce the pre-training technique.

\subsection{Architecture of the NIERT interpolator}\label{sec:arch_of_niert}


The neural interpolator in NIERT adopts the Transformer encoder framework; however, to suit the interpolation task, significant modifications and extensions were made in the embedding, Transformer and output layers, which are described in detail below.



{\bf Unified representations with masked tokens.} NIERT embeds both observed points and target points into the unified high-dimensional embedding space. As the position $x$ of a data point and its value $y$ are from different domains, we use two linear modules : $\mathrm{Linear}_x$ embeds the positions while $\mathrm{Linear}_y$ embeds the values. 

We feed the points and target points into the same embedding modules $\mathrm{Linear}_x,\mathrm{Linear}_y$. It should be noted that for target points, their values are absent when embedding as they are to be determined. In this case, we use a masked token as a substitute, which is embedded as a trainable parameter $\mathrm{MASK}_y$ as performed in BERT \cite{devlin2018bert}. This way, the interpolator processes both target points and observed points in a unified fashion. 

We concatenate the embeddings of position and value of a data point as the point's embedding, denoted as $h_i^0$, i.e.,  
\begin{equation*}
h_i^0 = \begin{cases}
\left[ \mathrm{Linear}_x (x_i), \mathrm{Linear}_y (y_i) \right], & \mathrm{if}  ~ (x_i, y_i) \in O \\
\left[ \mathrm{Linear}_x (x_i), \mathrm{MASK}_y \right], & \mathrm{if} ~ x_i \in T \end{cases}.
\end{equation*}

{\bf Partial self-attention mechanism.} 
NIERT feeds the embeddings of the points into a stack of $L$ Transformer layers, producing encodings of these points as result. Through the $l$-th layer, the encoding of the $i$-th point $h^l_i$ are transformed to $h^{l+1}_i$. Each Transformer layer contains two subsequent sub-layers, namely, a multi-head partial self-attention module, and a point-wise fully-connected network. These sub-layers are interlaced with residual connections and layer normalization between them.
 

 To avoid the unexpected interference of target points on observed points and target points themselves, NIERT replaces the original self-attention in the Transformer layer with {\it partial self-attention} mechanism.
As in vanilla self-attention \cite{vaswani2017attention}, the point embedding $h^l_i$ is firstly transformed to the query vector $q_i^l$, key vector $k_i^l$ and value vector $v_i^l$ by linear projection matrices $W_Q^l,~W_K^l,~W_V^l$, respectively. Then, the unnormalized attention score $e^l_{ij}$ can be computed by scaled dot production of the query $q_i^l$ and key $k_j^l$: 
\begin{align}
e^l_{ij} &= \frac{q_i^l\cdot k_j^l}{\sqrt{d_{k}}}
\end{align} 
where $d_k$ is the dimension of the query and key vectors. Unlike the vanilla self-attention layer, we mask out entries of attention weights between target points and all points by replacing these entries with $-\infty$:
\begin{equation*}
\hat{e}_{ij} = \begin{cases}
-\infty, & \mathrm{if}  ~ x_j \in T, \\
e_{ij}, &\mathrm{otherwise}
\end{cases}.
\end{equation*}
Then the attention score can be obtained by softmax normalization: \begin{align}
\alpha^l_{ij} =
\frac{\exp{(\hat{e}_{ij}^l)}}{\sum_{p}\exp{(\hat{e}_{i,p}^l)}} 
\end{align}
In this way, a masked entry $\alpha^l_{ij}$ is equal to $0$ when the $j$-th point is a target point, which is also illustrated in Fig.~\ref{fig:pipeline}. Then the layer output $h^{l+1}_{i}$ can be computed as follows:
\begin{align}
    h^{l+1}_{i} &= \mathrm{LayerNorm}(\tilde{v}^{l}_{i} + \mathrm{MLP}(\tilde{v}^{l}_{i})),  \\
    \text{where } \tilde{v}_{i}^{l} &= \mathrm{LayerNorm} \Big(v^{l}_{i} + \sum_{j} \alpha^l_{ij} v_{j}^{l} \Big).  \label{eq_pattn_layer}
\end{align}

Through partial self-attention, for each observed point $(x_j, y_j) \in O$, we maintain $\alpha^l_{ij}$ not masked; thus, NIERT can model the correlation between observed points and target points and the correlation among observed points themselves (Fig. \ref{fig:pipeline}).  Relatively, the correlation among observed points is easier to learn as these points have known values. In addition, this correlation can be transferred onto the target points, thus promoting learning representations of these data points. In contrast, by forcing the weight $\alpha^l_{ij}$ to be 0 for each target point $j$, we completely avoid the unexpected interference of target points on the other points. This design will eventually lead to the more robust representation of the target points and more accurate values will be estimated.


{\bf Estimating values for target points.} For each target point $i$, we estimate its value $\hat{y_{i}}$ through feeding its features at the final Transformer layer into a fully connected feed-forward network, i.e., 
\begin{equation*}
\hat{y_{i}} = \mathrm{MLP_{out}} (h_{i}^{L}). 
\end{equation*}
We calculate the error between the estimation and the corresponding ground-truth value, and compose the errors for all points into a loss function $\mathcal{L}$ to be minimized. 




\subsection{Pre-training NIERT interpolators}
\label{sec:pretrain}
We pre-train the NIERT interpolation model to obtain powerful pre-trained interpolators and apply them to various interpolation scenarios (Fig~\ref{fig:pretrain_pipeline}).
Since the dimension of the input and output of the function, $d_x$, $d_y$, which affects the architecture of the interpolator, may be different in various interpolation scenarios, we build a series of pre-trained interpolation models to accommodate scenarios with different function dimensions $(d_x,d_y)$.
The pre-trained NIERT in a certain $(d_x,d_y)$ configuration has strong generalization power to different scenarios with the same dimension configuration. For example, the pre-trained NIERT with $(2,1)$ dimension can be transferred into 3 different subsets of 2D temperature field reconstruction, i.e., HSink, ADlet and DSine, which are temperature field data for three quite different types of physical conditions\cite{chen2021machine}.

The key to the pre-training process is the construction of the pre-training task set. We propose a \textbf{math} function \textbf{i}nterpolation \textbf{t}ask set, namely, \textbf{Mathit}, for pre-training our model. These tasks for pre-training are sampled from synthesized mathematical functions (see Fig~\ref{fig:pretrain_pipeline}). These mathematical functions can be evaluated numerically with perfect accuracy and are suitable for sampling interpolation tasks to train highly accurate interpolators.
The configured synthesis process is flexible to generate functions of different dimensions, which are adequate for training interpolator families. Synthesizing data exhibits significant advantages over collecting data from the real world on large scale and at low costs. 

\begin{table}[h]
    \caption{Operators with corresponding sample weights used in generating mathematical functions for pre-training}
    \label{table:Mathit_symbols}
    \centering
    \begin{tabular}{ccccccccc} \toprule
        Operator & $+$  & $\times$ & $-$ & $\cdot^2$ & $\cdot^3$ & $\exp$ & $\sin$ & $\cos$ \\ \midrule
        Sample weight     &  10 & 10 & 5 & 4 & 2 & 4 & 4 & 4 \\ \bottomrule
    \end{tabular}
\end{table}

{\bf Synthesis of mathematical expressions. } 
 We borrow the work of \cite{lample2019deep,biggio2021neural} to randomly synthesize mathematical functions. A mathematical function is synthesized in the form of an expression tree, where the internal nodes are operators and the leaf nodes are independent variables and coefficients. Eight types of operators with different sampling weights are used in the random synthesis (Table~\ref{table:Mathit_symbols}), and each synthesized expression have at most 6 operators. We use these operators excluding `$\div$' or `$\log$', because it should be guaranteed that the synthesized functions are continuous so as to sample legitimate interpolation tasks.
Leaf nodes have a probability of $0.8$ to be independent variables and a probability of $0.2$ to be coefficients. The optional set of independent variables depends on the dimensionality of the scattered point in the task. For example, when the dimension of the scattered points $d_x$ is $3$, the independent variables are chosen with equal probability as $x_1$, $x_2$, $x_3$.
 For each dimension of function input $d_x$, we randomly synthesized one million of these expressions with coefficients and saved them for the subsequent sampling of the interpolation task. 

{\bf Sampling pre-training interpolation tasks. } 
At training time, the coefficients of each expression are randomly sampled from $[-2,2]$, such that the functions seen by the model are all different. Once the coefficients in an expression are determined, the expression can be numerically evaluated as a mathematical function. Next, an interpolation task is sampled from each evaluated function. We randomly sample $256$ scattered points from within $[-1,1]^{d_x}$, and compute the values of the function on these points. The function values of the scattered points can be very large or small, which may lead to exploding or vanishing gradients during training, so we normalize the values of each set of scattered points. To construct an interpolation task, we divide these $256$ points into $n$ observed points and $256-n$ target points randomly. Here, we randomly set $n$ to an integer in $[10,50]$, which induces the generalizability of our model to a variable number of observed points. In this way, an interpolation task for pre-training is sampled.

{\bf Dimensional augmentation technique. } To synthesize mathematical functions with output dimension greater than $1$ ($d_y>1$), we use the dimensional augmentation technique, which randomly takes $d_y$ functions from the mathematical function set with independent variable dimension $d_x$ and concatenates them together. We do this based on the simple fact that functions of multiple dependent variables can be directly expressed as a combination of multiple functions of one dependent variable. For example, a two-dimensional velocity field can be expressed as $[v_x(x,y), v_y(x,y)]$. 

{\bf Pre-training and fine-tuning. } We pre-train this series of NIERT interpolators separately for different $(d_x, d_y)$ configurations. 
For a specific $(d_x,d_y)$ configuration, the pre-trained NIERT interpolator can be transferred to different interpolation scenarios of the same dimension configuration.
After pre-training, we obtain pre-trained NIERT interpolators with general interpolation capabilities that can be fine-tuned on a domain-specific set of interpolation tasks.


\begin{table*}[hbt!]
\caption{ Statistics of the interpolation tasks used for training in each dataset.}
\label{table:dataset_all}
\centering
\begin{tabular}{lllllll} \toprule
        \multirow{1}{*} [-0pt] { \makecell{Dataset}} & Subset tag & $x$'s dimenstion $d_x$ & $y$'s dimenstion $d_y$ &   $\#$All points $N$ & $\#$Observed points $n$ & $\#$Target points $m$ \\ \midrule
        \multirow{1}{*} [-0pt] { \makecell{Mathit}}  & \multicolumn{1}{l}{1D/2D/3D/4D} & \multicolumn{1}{l}{$1/2/3/4$} &  \multicolumn{1}{l}{1} & \multicolumn{1}{l}{$256$} & \multicolumn{1}{l}{$[10,50]$} & $N-n$ \\  
        \multirow{1}{*} [-0pt] { \makecell{TFRD}}     & ADlet/DSine/HSink & $2$ & $1$  &  $40000$  & $37$ & $N-n$  \\  
        \multirow{1}{*} [-0pt] { \makecell{PTV}}     & - & $2$ & $2$  &  $[2291, 5899]$  & $512$ & $N-n$  \\  
        \multirow{1}{*} [-0pt] { \makecell{PhysioNet}}  & \multicolumn{1}{l}{-} & 1 & 41 & \multicolumn{1}{l}{$[20,190]$} & $0.5N / 0.7N / 0.9N$ & $N-n$ \\   \bottomrule
    \end{tabular}
\end{table*}

\section{Experiment setting}
\label{sec:exp_setting}

In this section, we present the description of datasets used for our experiments, the baselines used for comparison, and the experimental details.

\subsection{Datasets}

We evaluate our approach on four representative datasets, including two synthetic datasets and two real-world datasets. These datasets are representatives of interpolation tasks in various application fields: Mathit for mathematical function interpolation, TFRD \cite{chen2021machine} for temperature field reconstruction, PTV \cite{sciacchitano2015collaborative} for two-dimensional particle tracking velocimetry, PhysioNet \cite{silva2012predicting} for time-series data interpolation. We introduce each dataset in detail as follows.

{\bf 1) Synthetic dataset I: Mathit}, which denotes the pre-trained dataset we constructed (described in detail in Section~\ref{sec:pretrain}), is used to test the performance for mathematical function interpolation. We evaluate the interpolation accuracy of NIERT and existing approaches on functions of various independent variable dimensions $d_x$, including $1D$, $2D$, $3D$ and $4D$. 
In addition to the pre-trained dataset, we synthesized a test set including $12,000$ interpolation tasks for each independent variable dimension of functions. 
To avoid the intersection between training and test sets, we remove the mathematical expressions appearing in both sets.

{\bf 2) Synthetic dataset II: TFRD} is used to test the performance for 2D temperature field reconstruction. It consists of three sub-datasets with large differences: HSink, ADlet, and DSine, each of which simulates the temperature field of mechanical devices using a specific kind of heat generation and boundary conditions\cite{chen2021machine}. Each reconstruction task in each subset has $200\times200$ regular grid points that represent the temperature field in a $0.1m\times 0.1m$ square area. Among these points, $37$ scattered points have their temperate known and are used as observed points. The other points are used as target points. In each subset, we have a total of $10,000$ training instances and $10,000$ test instances. We follow to use $L_1$-form loss function for this dataset as performed in \cite{chen2021machine} for fair a comparison.
 
{\bf 3) Real-world dataset I: PTV} is a scattered data set where the scattered points are particles with velocities in the flow field. PTV is used to test the performance for particle tracking velocimetry, i.e., the two-dimensional velocity field reconstruction from a finite number of observed particles with velocities. These data are extracted from the raw images of the laminar jet experiment scenes taken by \cite{sciacchitano2015collaborative}. There are $1200$ raw frames in total corresponding to $1200$ velocity fields, and each frame extracts a set of scattered points with velocity, which have at most close to $6000$ points and at least $2000$ points. For each set of scattered data, we randomly take $512$ points as observed points and the remaining points as target points to construct an interpolation task. We randomly select a quarter of the tasks as the test set and the rest as the training set.

{\bf 4) Real-world dataset II: PhysioNet} excerpted from the PhysioNet Challenge 2012 \cite{silva2012predicting}, is used to test the performance of NIERT for irregularly-sampled time series interpolation. This real-world dataset was collected from intensive care unit (ICU) records. 
Each instance consists of multiple points, each point represents a measurement of a patient at a specific time, and each measurement contains up to $41$ physiological indices. Following the study in \cite{shukla2021multitime}, we randomly divided the points into observed points and target points, then trained and evaluated interpolation models using the acquired interpolation task sets. We set the ratio of observed points at three levels, i.e., $50\%$, $70\%$, and $90\%$ for adequate comparisons. We randomly divided the $8,000$ instances of PhysioNet into the training set and test set with a ratio of $4:1$.
Since these time series as functions have a dependent variable of dimension $41$, to evaluate the effectiveness of the pre-training on this dataset, we additionally pre-trained a NIERT model suitable for such functions using the dimensional augmentation technique.

We list the statistics of all the above datasets in Table~\ref{table:dataset_all}.
In practice, for each instance in the training set, i.e., $N$ scattered points from a certain function, we randomly select $n$ of them as observed points and the remaining $N-n$ as target points. For the Mathit dataset, we randomly select a fixed range of numbers as the number of observed points (e.g. $[5,50]$ for Mathit dataset). For the real-world PhysioNet dataset, which has a variable number of scattered points for its instances, distributed in the range of $[20,190]$, we, therefore, select observed points from all points at a fixed rate (say 50\%,70\% or 90\%).

\subsection{Baselines}
All baselines can be divided into two categories: 1) traditional interpolation algorithms and 2) learning-based models.


\subsubsection*{Traditional Interpolation Algorithms.}

\noindent {\bf Radial basis function (RBF) interpolation} is one of the most commonly-used scattered data interpolation methods. It adopts a specific type of radial basis function on observed points and uses their linear combination to represent the target function. We use the RBF interpolation implementation in SciPy \cite{2020SciPy-NMeth} and multiquadric function as the basis function type for the experiments.

\noindent {\bf MIR} is another multivariate interpolation and regression method for scattered data sets proposed by \cite{wang2010high}. MIR represents the approximation error with Taylor expansions and minimizes the approximation error to find the basis functions.


\subsubsection*{Learning Based Models.}

\noindent {\bf Conditional neural processes (CNPs)} proposed by \cite{pmlr-v80-garnelo18a} is a neural model composed of MLPs, which is able to learn to predict distributions of functions given observed points.

\noindent {\bf Attentive neural processes (ANPs)} \cite{kim2018attentive} leverages the attention mechanism in CNPs and improves the prediction performance.

\noindent {\bf Bootstrapping attentive neural processes (BANPs)} ~ \cite{NEURIPS2020_492114f6} employs the bootstrap technique to further improve the performance of ANPs.

\noindent {\bf TFR-Transformer} , proposed in \cite{chen2021machine}, uses Transformer \cite{kim2018attentive} in 2-dimensional temperature field reconstruction using scattered observations.

\noindent {\bf RNN-VAE} is a VAE-based model where the encoder and decoder are standard RNN models. Gated Recurrent Unit (GRU) \cite{chung2014empirical} module is configured as the recurrent network.

\noindent {\bf L-ODE-RNN} refers to the latent neural ODE model where the encoder is an RNN and the decoder is a neural ODE proposed in \cite{chen2018neuralode}.

\noindent {\bf L-ODE-ODE} refers to the model where the encoder is an ODE-RNN \cite{NEURIPS2019_42a6845a} and the decoder is a neural ODE.

\noindent {\bf mTAND-Full} \cite{shukla2021multitime} performs time attention mechanism and Bidirectional RNNs to encode temporal features  and interpolate irregular-sampled time series data.

\begin{table}[hbt!]
\caption{Hyper-parameters of NIERT in experiments on the four datasets.}
\label{table:params_all}
\centering
\begin{tabular}{lrrrr} \toprule
    \multirow{2}{*} [-2pt] { \makecell{Parameter name} } 
             &   \multicolumn{4}{c}{Parameter value in experiments on} \\ \cmidrule{2-5}
         & \multicolumn{1}{c}{Mathit} & \multicolumn{1}{c}{TFRD} & \multicolumn{1}{c}{\quad PIV} & \multicolumn{1}{c}{PhysioNet} \\ \midrule
        $x$'s embedding dimension  & $16\times d_x$  & $32$ & $32$ & $16$ \\
        $y$'s embedding dimension  & $16$  & $16$ & $32$ & $592$ \\  
        Number of layers   &  $6$ & $3$ & $3$ & $3$ \\
        Number of heads    &  $8$ & $4$ & $4$ & $4$ \\
        Hidden dimension       & $512$  & $128$ & $128$ & $128$ \\
        \bottomrule
    \end{tabular}
\end{table}

\begin{table*}[!t]
  \begin{minipage}{0.48\textwidth}
    \scriptsize
    \caption{Interpolation accuracy on Mathit  dataset.
    }
    \label{table:Mathit}
    \centering
    \resizebox{8.4cm}{!}{
    \begin{tabular}{crrrrr} \toprule
    \multirow{2}{*} [-2pt] { \makecell{Interpolation\\approach} }
             &   \multicolumn{4}{c}{MSE ($\times 10^{-5}$) on Mathit test set} \\ \cmidrule{2-5}
         & \multicolumn{1}{c}{1D} & \multicolumn{1}{c}{2D} & \multicolumn{1}{c}{3D} & \multicolumn{1}{c}{4D} \\ \midrule
        RBF        & 215.439 & 347.060 & 443.094 & 327.775  \\
        MIR         & 67.281  & 274.601 & 448.933 & 342.997  \\
        CNP  & 67.176  & 248.668 & 392.348 & 314.311  \\
        ANP   & 34.558       & 140.005 & 206.699 & 164.751  \\
        BANP   & 14.913       & 84.187 & 143.518 & 140.288  \\
        TFR-transformer & 15.556  & 58.569  & 99.986  & 90.579  \\
        NIERT  & \textbf{8.964} & \textbf{45.319} & \textbf{77.664} & \textbf{72.025}  \\ \bottomrule
    \end{tabular} }
  \end{minipage}
  \hfill
  \begin{minipage}{0.48\textwidth}
    \scriptsize
    \caption{Interpolation accuracy on TFRD dataset.}
    \label{table:TFRD}
    \centering
    \resizebox{8.5cm}{!}{
    \begin{tabular}{crrr} \toprule
        \multirow{2}{*} [-2pt] { \makecell{Interpolation\\approach} } &   \multicolumn{3}{c}{MAE ($\times 10^{-3}$) on TFRD test set } \\ \cmidrule{2-4}
                        & \hspace*{1.5mm} HSink & \hspace*{1.5mm} ADlet & \hspace*{1.5mm} DSine \\ \midrule
        CNP             & 204.351     & 91.782     & 92.456 \\
        ANP             & 164.491       & 54.684     & 58.589 \\
        BANP            & 59.728         & 28.671     & 19.107 \\
        TFR-transformer & 64.987    & 27.074     & 29.961 \\
        NIERT           & 23.519    & 3.473      & 8.785 \\
        NIERT w/ pretraining  & \textbf{15.076} & \textbf{1.897} & \textbf{4.912} \\ \bottomrule
    \end{tabular}}
  \end{minipage}
  \end{table*}

\begin{table*}[!t]  
  \begin{minipage}{0.44\textwidth}
  
    \scriptsize
    \caption{Interpolation accuracy on PTV dataset.}
    \label{table:PTV}
    \centering
    \resizebox{6.2cm}{!}{\begin{tabular}{cc} \toprule
        \multirow{2}{*} [-2pt] { \makecell{Interpolation\\approach} } &   \multicolumn{1}{c}{Evaluation criteria} \\ \cmidrule{2-2}
                   & MSE  ($\times 10^{-3}$) \\ \midrule
        CNP             &     137.573   \\
        ANP             &     32.111  \\
        BANP             &    33.585    \\
        TFR-transformer     & 17.125 \\
        NIERT          &      5.167   \\
        NIERT w/ pretraining  & \textbf{1.954}  \\ \bottomrule
    \end{tabular}
    }
  \end{minipage}
  \hfill
  \begin{minipage}{0.54\textwidth}
    \scriptsize
    \caption{Interpolation accuracy (MSE, $\times 10^{-3}$) on PhysioNet dataset.}
    \label{table:PhysioNet}
    \centering
    \resizebox{10cm}{!}{\begin{tabular}{crrr} \toprule
        \multirow{2}{*} [-2pt] { \makecell{Interpolation\\approach} } &   \multicolumn{3}{c}{Ratio of observed points} \\ \cmidrule{2-4}
        & \multicolumn{1}{c}{50\%}  & \multicolumn{1}{c}{70\%} & \multicolumn{1}{c}{90\%} \\  \midrule
        RNN-VAE     & 13.418$\pm$0.008   & 11.887$\pm$0.005   & 11.470$\pm$0.006 \\
        L-ODE-RNN   & 8.132$\pm$0.020   & 8.171$\pm$0.030   & 8.402$\pm$0.022 \\
        L-ODE-ODE   & 6.721$\pm$0.109    & 6.798$\pm$0.143    & 7.142$\pm$0.066 \\
        mTAND-Full  & 4.139$\pm$0.029    & 4.157$\pm$0.053   & 4.798$\pm$0.036 \\
        NIERT     & 2.868$\pm$0.021    & 2.656$\pm$0.041    & 2.709$\pm$0.157 \\
        NIERT w/ pretraining & \textbf{2.831$\pm$0.021} & \textbf{2.641$\pm$0.052} & \textbf{2.596$\pm$0.159} \\ \bottomrule
    \end{tabular}
    }
    \end{minipage}
\end{table*}

\subsection{Experimental details}

\textbf{Model hyper-parameters.} Table~\ref{table:params_all} lists the hyper-parameters of NIERT in the experiments on the four representative datasets, including Mathit, TFRD, PIV and PhysioNet.
To fairly evaluate the performance of the pre-trained models, the hyperparameters of the pre-trained models we used on the TFRD, PIV and PhysioNet datasets were aligned with those of NIERT.
In addition, we configured the TFR-transformer with the same hyperparameter settings for a fair comparison.
For hyperparameter settings of other baselines, we follow the settings in the work of \cite{chen2021machine} and \cite{shukla2021multitime}.

\textbf{Evaluation metrics.} For the Mathit, PTV, and PhysioNet datasets, we used the mean square error (MSE) of interpolation on the target points to evaluate the interpolation accuracy of the model. For the TFRD dataset, we follow the work of \cite{chen2021machine} and used the mean absolute error (MAE) on the target points to evaluate the accuracy of the temperature field reconstruction. 

\textbf{Training details.} We used MSE as the loss function for training on Mathit , PTV, and PhysioNet datasets, and MAE as the loss function for training on TFRD dataset, which is consistent with the respective evaluation metric settings. We used Adam for parameter optimization. For pre-training on the Mathit dataset, we sampled mini-batches of size 150 and used a learning rate of $10^{-4}$, no schedules, and trained for 160 epochs. For training or fine-tuning on the TFRD, PTV and PhysioNet dataset, we sampled mini-batches of size 5, 32 and 4 respectively. The models were trained on these three datasets using a learning rate of $10^{-4}$, and the learning rate was decayed by a factor of $0.97$ at the end of each epoch.

\section{Results}

In this section, we present experimental results and analysis, including interpolation accuracy on synthetic and real-world datasets, individual case studies of interpolation, interpretability analysis of attentions, and results of ablation experiments. These results demonstrate the effectiveness and superiority of our approach. We present more results in the Supplementary Material.

\begin{figure}[h]
\centering
\subfloat[On Mathit-1D test set]{\includegraphics[width=2.4in]{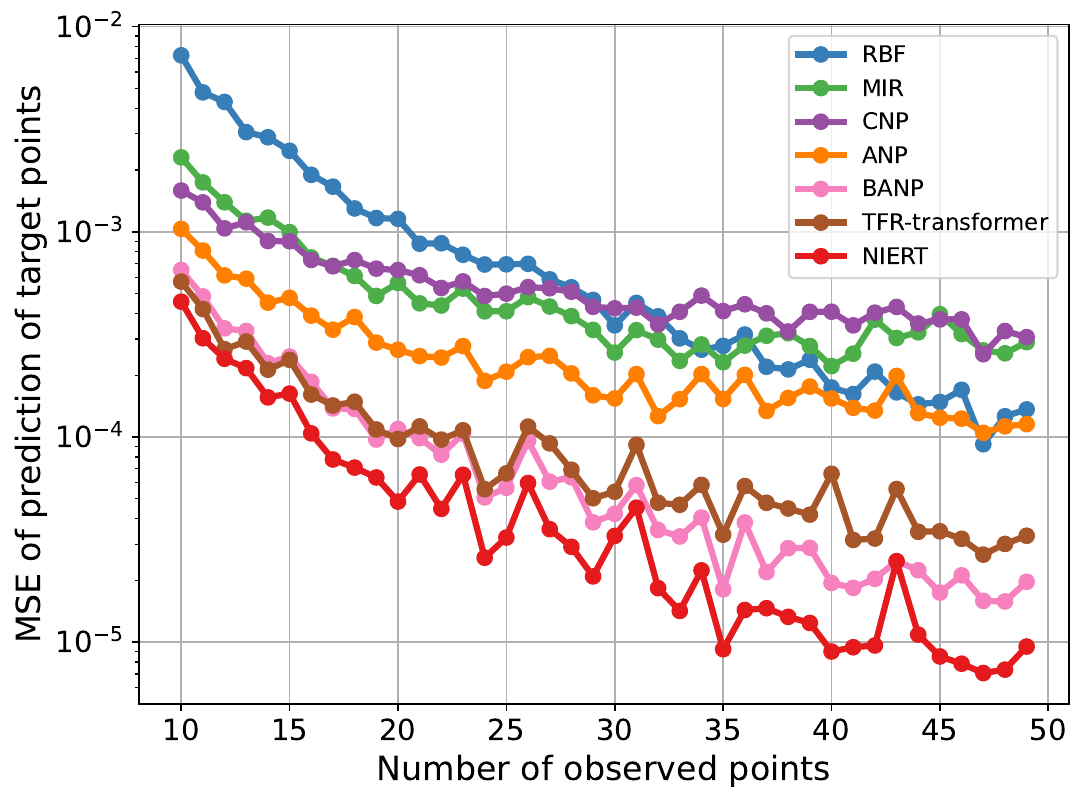}%
\label{fig:MSE_v_PN_1}}
\vfil
\subfloat[On Mathit-2D test set]{\includegraphics[width=2.4in]{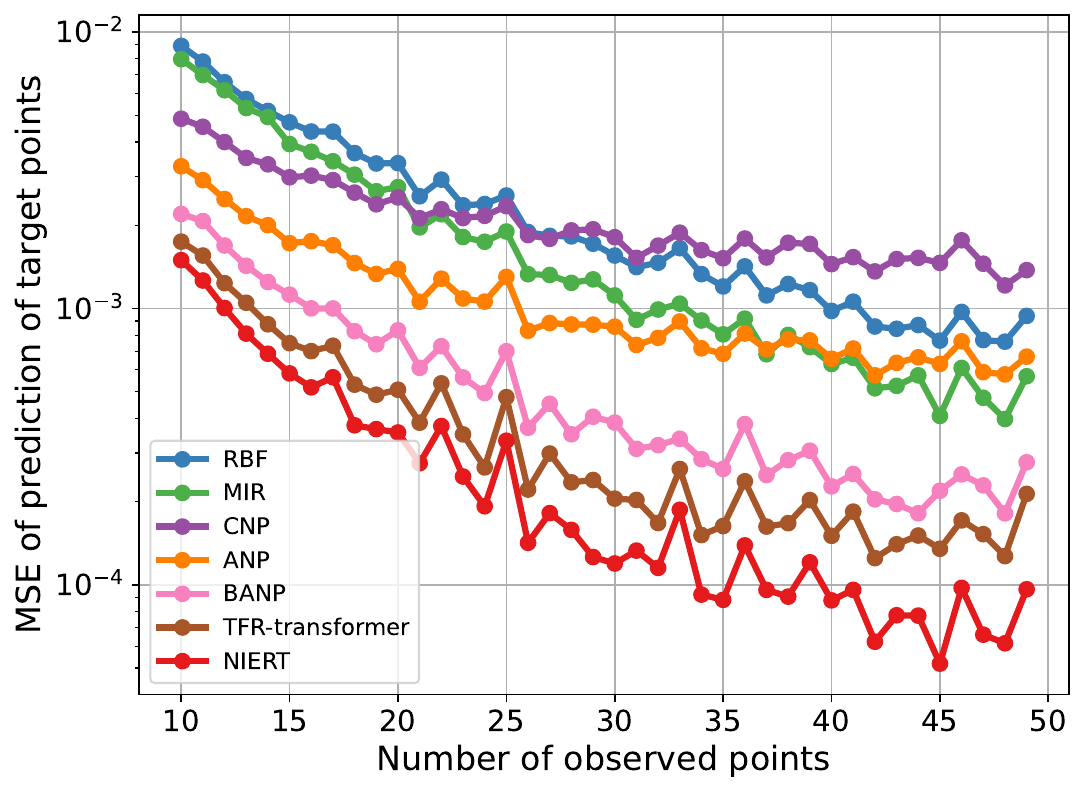}%
\label{fig:MSE_v_PN_2}}
\caption{The relationship between the interpolation accuracy and the number of observed points. Here we use the instances in the Mathit-1D and Mathit-2D test dataset as representatives.}
\label{fig:MSE_v_PN}
\end{figure}

\begin{figure*}[!htb]
    \centering
    {\includegraphics[width=13cm]{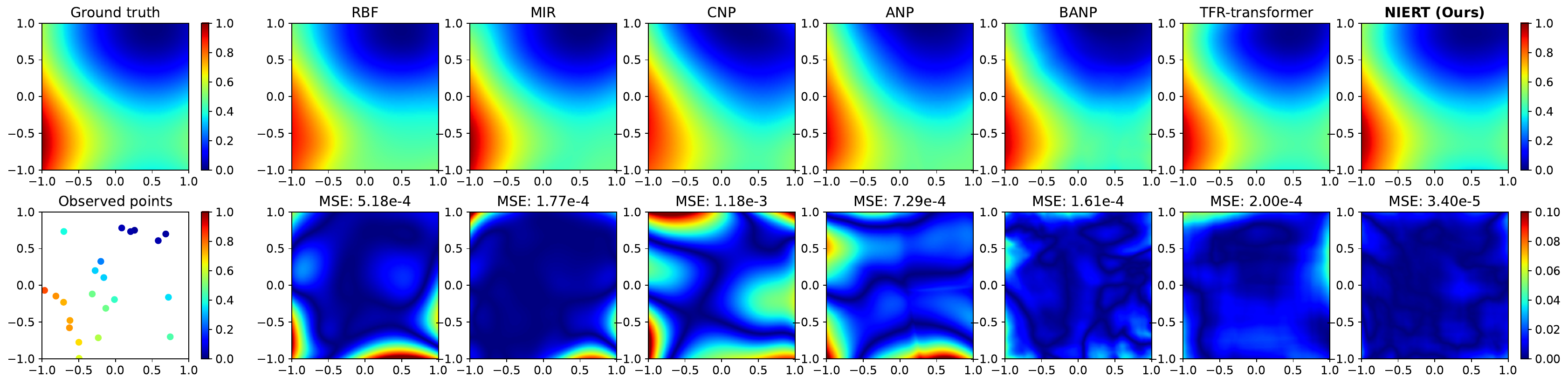}}\\
    \caption{An example of 2D interpolation task extracted from Mathit test set. The up-left figure shows the ground-truth function 
    $f(x_1,x_2)=  0.65 \cos\left(1.232\sin(x_1 + 0.636)\right) 0.25 x_2^2 - 0.25 x_2 + 0.156$
    while the bottom-left figure shows the 22 observed points. The interpolation functions reported by NIERT and the existing approaches are listed on the top row while their error maps {(their differences with the ground truth)} are listed below.}
    \label{fig:2d_vis}
\end{figure*}

\begin{figure*}[!h]
    \centering
    \includegraphics[width=12cm]{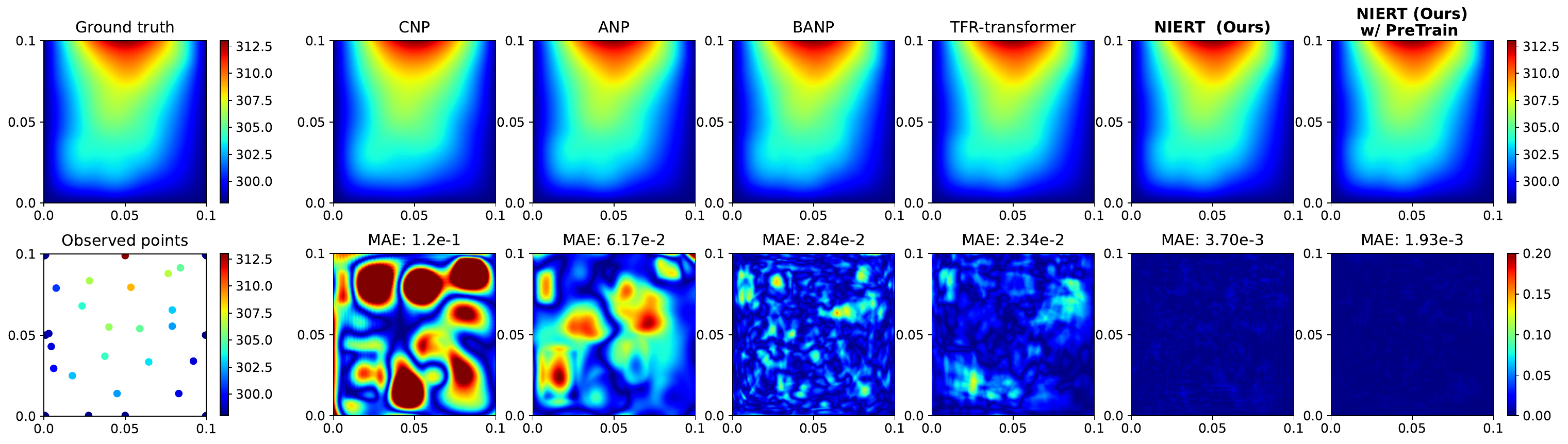}
        \caption{An example of a temperature field reconstruction task extracted from TFRD-ADlet test set. The up-left figure shows the ground-truth temperature field while the bottom-left figure shows the 32 observed points. The reconstructed results reported by NIERT and the existing approaches are listed on the top row their error maps {(their differences with the ground truth)} are listed below.}
        \label{fig:tfrd_vis}
\end{figure*}

\subsection{Interpolation accuracy on synthetic and real-world datasets}

For each instance of the test dataset, we applied the trained NIERT and the existing approaches to estimate values for target points. We calculate the mean of errors between estimation and ground truth as interpolation accuracy.

\textbf{Accuracy on Mathit datasets.} As shown in Table \ref{table:Mathit}, on the 1D Mathit test set, RBF shows the largest interpolation error (MSE: 215.439). MIR, another approach using explicit basis functions, also shows a high interpolation error of 67.281. In contrast, BANP and TFR-transformer, which use neural networks to learn interpolation, show relatively lower errors (MSE: 14.913, 15.556). Compared with these approaches, our NIERT approach achieves the best interpolation accuracy (MSE: \textbf{8.964} \textit{v.s.} 14.913). Table \ref{table:Mathit} also demonstrates the significant advantage of NIERT over the existing approach on the 2D, 3D, and 4D instances.

To examine in depth the interpolation accuracy, we further divide test instances into subsets according to the number of observed points. As shown in Fig~\ref{fig:MSE_v_PN_1} and Fig~\ref{fig:MSE_v_PN_2}, as the number of observed points increases, the interpolation error decreases as expected. In addition, the relative advantages of these approaches vary with the number of observed points, e.g., CNP is better than RBF and MIR initially but finally becomes worse as the number of observed points increases. Among all approaches, NIERT stably shows the best performance over all test subsets, regardless of the number of observed points.

\textbf{Accuracy on TFRD, PTV and PhysioNet datasets.} 
As shown in Table \ref{table:TFRD}, CNP, although employing the neural network technique, still performs poorly on TFRD datasets. In contrast, NIERT achieves much lower interpolation error (\textbf{23.519} \textit{v.s.} 204.351, \textbf{3.4739} \textit{v.s.} 91.782 and \textbf{8.785} \textit{v.s.} 92.456 on HSink, ADlet and DSink subsets, respectively), which are also significantly lower than ANP, BANP and TFR-transformer. Moreover, when enhanced with the pre-training technique, NIERT can consistently further decrease its interpolation MAE by 35\%, 45\% and 44\% relatively on these 3 subsets, respectively, which verifies the strong generalization power of our pre-trained NIERT interpolator.

Table~\ref{table:PTV} shows that NIERT also outperforms existing methods on the PTV dataset, for example, the average MSE ($\times 10^{-3}$) of NIERT is significantly lower than all other methods (\textbf{1.954} \textit{v.s.} 17.125). Similarly, the pre-trained NIERT shows better performance. This demonstrates the effectiveness of our NIERT and pre-training method in the 2D particle tracking velocimetry scenario.

Table~\ref{table:PhysioNet} suggests that on the PhysioNet dataset for irregularly-sampled time series interpolation, NIERT also outperforms the existing approaches, e.g., when controlling the ratio of observed points to be 50\%, the interpolation error of NIERT significantly lower than other approaches (\textbf{2.868} \textit{v.s.} 4.139). Again, NIERT with the pre-training technique shows better performance. The advantages of NIERT hold across various settings of the ratio of the observed points.

Taken together, these results demonstrate the power of NIERT for numerical interpolation in multiple application fields, including mathematical function interpolation, temperature field reconstruction, particle tracking velocimetry, and interpolating irregularly-sampled time-series data. These results also show that the proposed pre-training technique can give NIERT interpolators strong generalization capabilities.

\subsection{Case studies of interpolation results}
To further understand the advantages of NIERT, we carried out case studies through visualizing the observed points, the reconstructed interpolation functions and the interpolation errors in this subsection. We have placed an example of Mathit-2D and an example of TFRD-ADlet in Fig.~\ref{fig:2d_vis} and Fig.~\ref{fig:tfrd_vis}, respectively. More visualized cases are provided in the Supplementary material.

Fig.~\ref{fig:2d_vis} shows a 2D instance in the Mathit test set. As illustrated, RBF performs poorly in the application scenario with sparse observed data. In addition, MIR and ANP, cannot accurately predict values for the target points that fall out of the range restricted by observed points. The CNP approach can only learn the rough trend stated by the observed points, thus leading to significant errors. In contrast, BANP, TFR-transformer and NIERT can accurately estimate values for target points within a considerably large range, and compared with BANP and TFR-transformer, NIERT can produce more accurate results.

Fig.~\ref{fig:tfrd_vis} shows an instance of temperature field reconstruction extracted from TFRD-ADlet. From this figure, we can observe that NIERT and pre-trained NIERT have significantly lower interpolation errors over the entire region than existing methods. When using the pre-training technique, NIERT further improves its interpolation accuracy in the whole area (with fewer light-colored blobs than non-pre-trained NIERT).

\subsection{Interpretability analysis} \label{subsec_contrib}
As demonstrated in Section~\ref{sec:method}, the partial self-attention mechanism has a good interpretability for the scattered interpolation task, where the attention score indicates the correlation between observed points and the correlation between observed and target points.
To analyze the interpretability, we visualized the attention scores in the last attention layer, i.e., $\alpha^L_{ij}$ in Eq. (\ref{eq_pattn_layer}), of each observed point to all target points. Here, we take a Mathit-2D interpolation task and set the target points to be points from a $128\times128$ uniform grid covering the $[-1,1]^2$ domain, such that these attention weights provide an intuitive description of the contribution by observed points. 

\begin{figure}[hbt!]
    \centering
    \includegraphics[width=7cm]{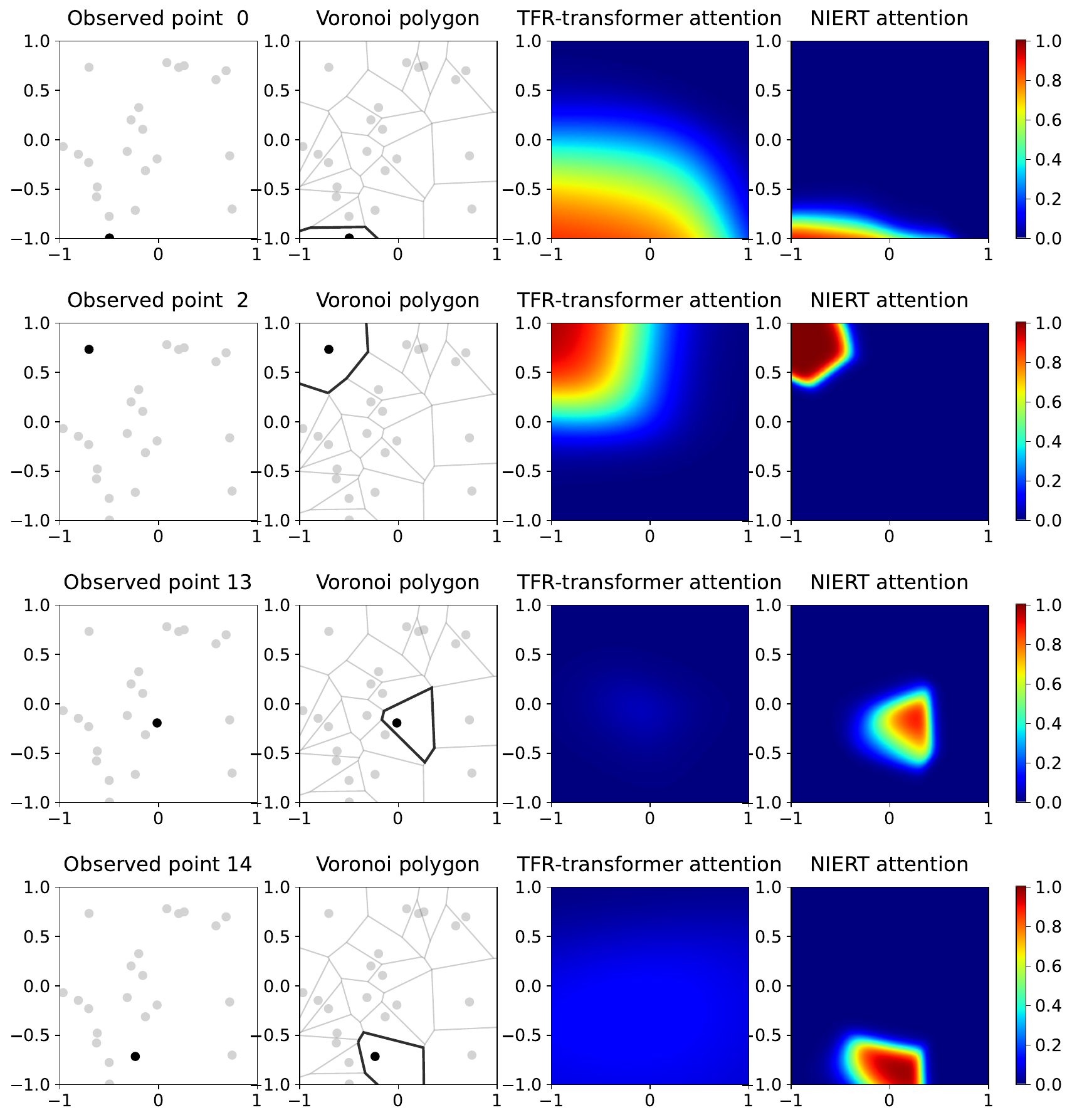}
    \caption{Contributions of certain observed points for interpolation. 
    We select 4 observed points and extract their attention weights from the final attention layer of NIERT and TFR-transformer. For analysis, we also visualized the Voronoi polygons \cite{aurenhammer1991voronoi} of the observed points. The contributions by the other 18 observed points are shown in Section~\ref{subsec:contrib} in the Supplementary material. }%
    \label{fig:attention}%
\end{figure}

As shown in Fig.~\ref{fig:attention}, we compared the attention weights from the final attention layer of NIERT and TFR-transformer. As a reference, we also present the Voronoi polygon \cite{aurenhammer1991voronoi} corresponding to each observed point, 
which is exactly the region contributed by each observed point in the nearest neighbor interpolation algorithm.
Due to the continuity of the target function, the values of the closer points are close to each other, thus there are strong correlations between such scattered points. Therefore, an ideal highly accurate interpolator should make the observed points tend to affect the target points closer to it rather than more distant ones.
Interestingly, we can find that using NIERT, each observed point contributes much more locally and targetedly on an area similar to its Voronoi polygon, which presents a strong interpretability.
Moreover, the areas and intensities contributed by each observed point do not overlap exactly with the Voronoi polygon, but are soft and more adaptive.
In contrast, when using TFR-transformer, the contributions by observed points are considerably imprecise and imbalanced. 
These results demonstrate that NIERT has better interpretability and can model the correlation between observed and target points more effectively and accurately. 


\subsection{Ablation study}

\textbf{The effects of partial self-attention.} For a specific interpolation task, the interpolation function is determined by the observed points only.
To investigate the effects of partial self-attention in avoiding interference of target points, we evaluated NIERT on the test sets with various numbers of target points. Here, we compared NIERT with its two variants, one using vanilla self-attention, and the other using partial self-attention with the target points' correlation added. Both of them were trained using the same training sets (the number of target points varies within $[206, 246]$).

\begin{figure}[!t]
\centering
\includegraphics[width=2.4in]{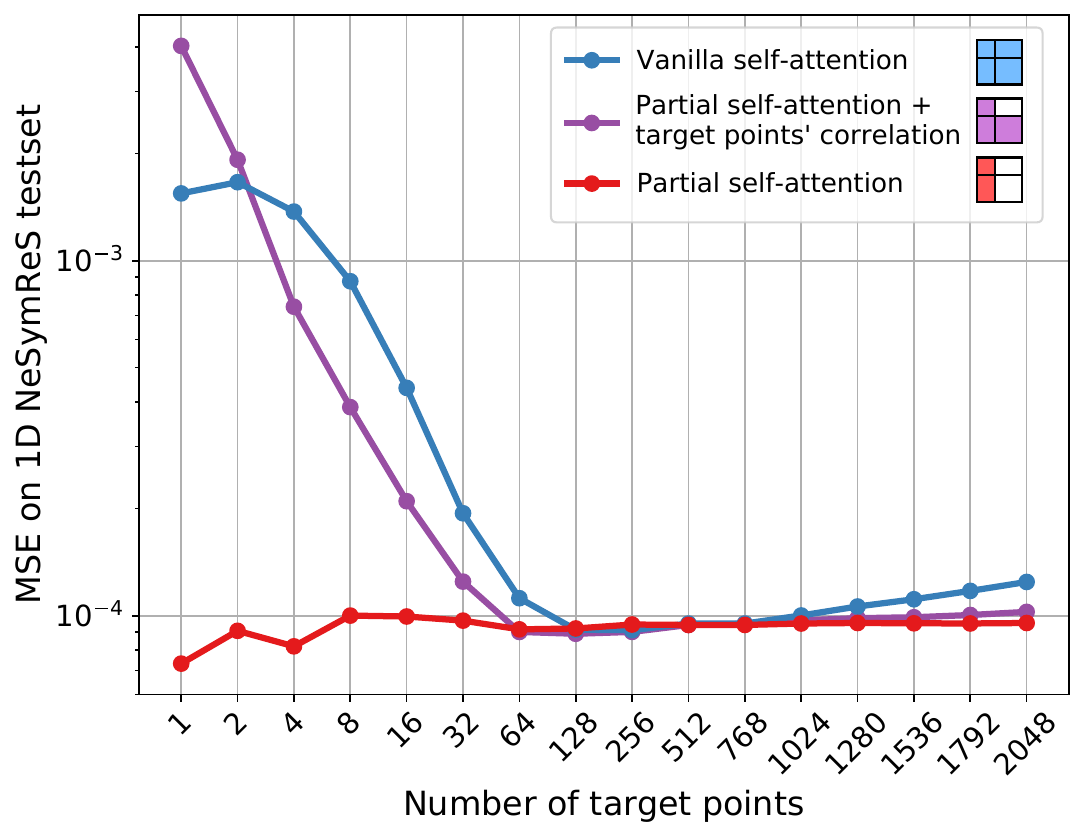}
\caption{The robustness of NIERT to the number of target points. Here, NIERT uses three attention mechanisms trained on the Mathit-1D dataset}
\label{fig:ablation_partial}
\end{figure}

As illustrated by Fig.~\ref{fig:ablation_partial},  the two NIERT variants show poor performance for the tasks with few target points, say less than 64 target points or more than 768 target points. In contrast, NIERT, which uses partial self-attention,  always performs stably without significant changes in accuracy. The results clearly demonstrate that the partial self-attention mechanism allows NIERT to be free from the unexpected effects of the target points, which enhances the robustness of interpolation.

\begin{figure}[h]
\centering
\subfloat[On TFRD-ADlet dataset]{\includegraphics[width=2.4in]{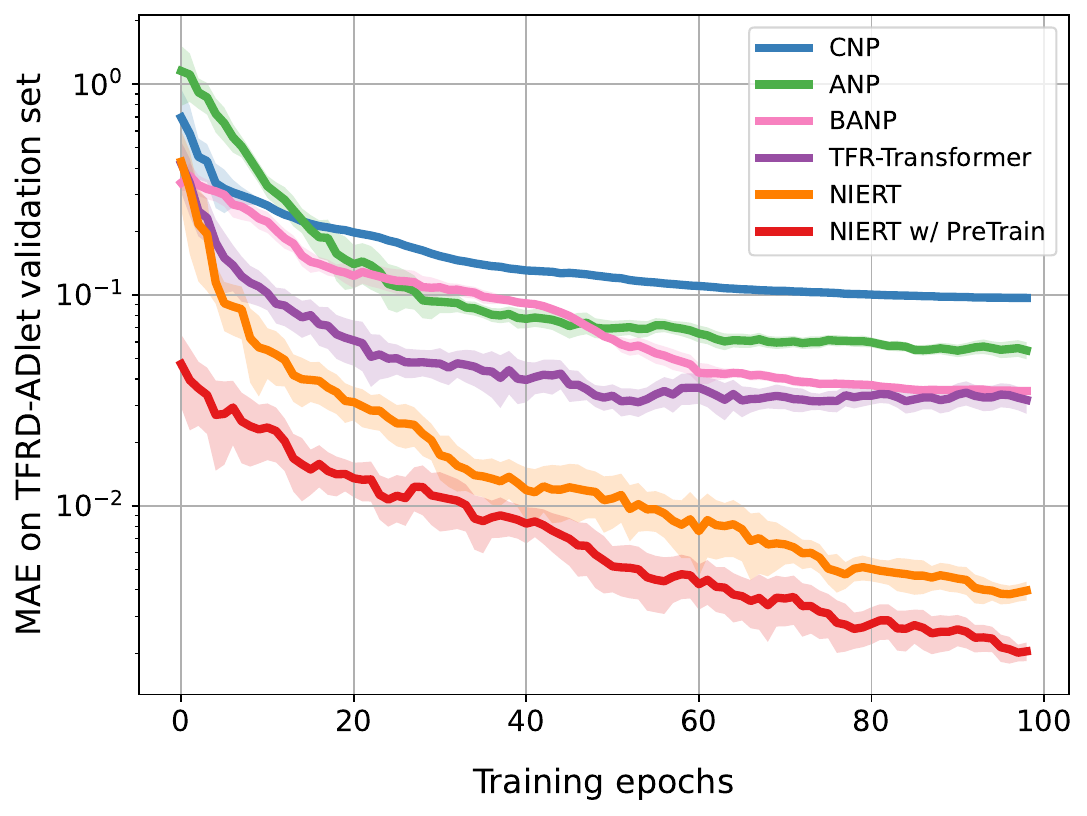}%
\label{fig:loss_iter_TFR}}
\vspace{0.1in}
\subfloat[On PTV dataset]{\includegraphics[width=2.4in]{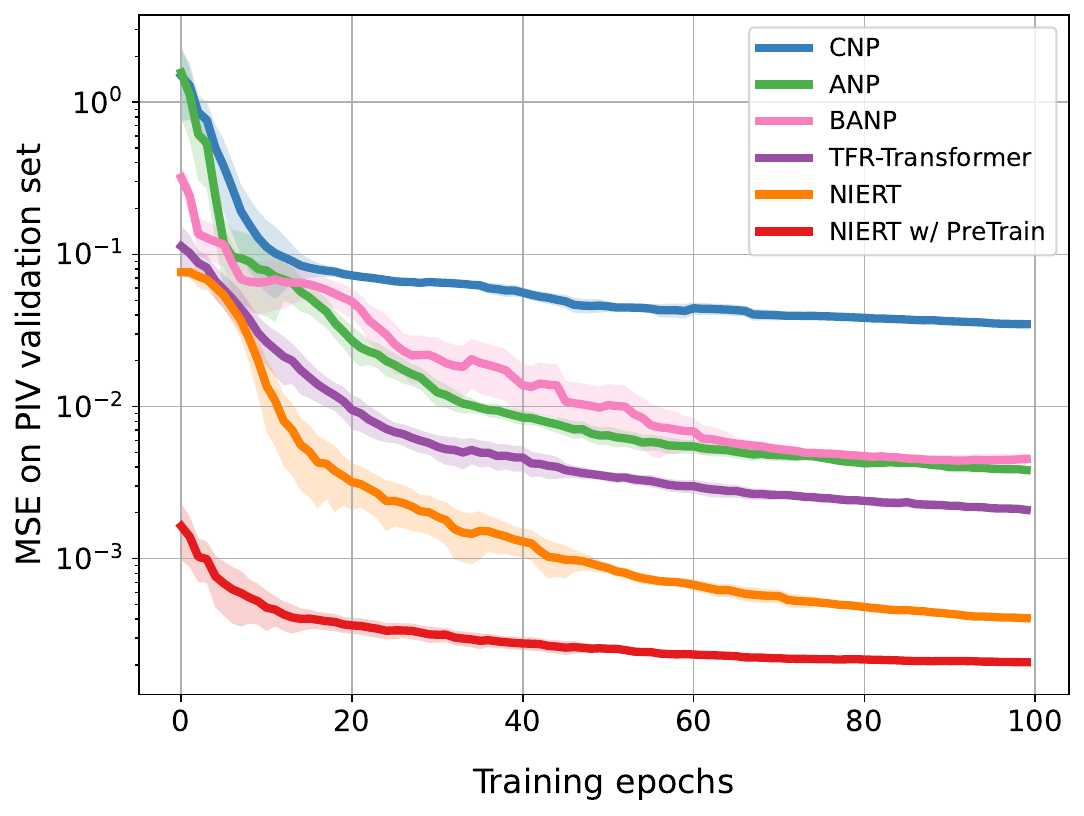}%
\label{fig:loss_iter_PTV}}
\caption{The convergence of NIERT, NIERT with pre-training, and the existing approaches. Here, these models are trained on TFRD-ADlet and PTV dataset.}
\label{fig:loss_iter}
\end{figure}

\textbf{The effects of pre-training technique.} 
To investigate the effects of the pre-training technique, we show in Fig~\ref{fig:loss_iter} the training process of two versions of NIERT, one without the pre-training technique, and the other enhanced with pre-training. As depicted by Fig~\ref{fig:loss_iter_TFR}, on TFRD-ADlet dataset, even only after the first epoch, the pre-trained NIERT shows a sufficiently high interpolation accuracy, which is comparable with the fully-trained BANP and TFR-transformer. On PIV dataset (Fig~\ref{fig:loss_iter_PTV}), the pre-trained NIERT has exceeded the final interpolation accuracy of all existing baselines even after the first epoch of fine-tuning. Moreover, the pre-trained NIERT ends up with the smallest interpolation errors on both datasets, which are nearly half of the error of the NIERT without pre-training.

These results suggest that the pre-training technique gives the NIERT interpolator powerful interpolation generalization capabilities. On one hand, it significantly accelerates the convergence and reduces the cost of transfer learning. On the other hand, it improves the final interpolation performance of NIERT.


\section{Discussion: connections to traditional approach}
\label{sec:discussion}

\label{subsec_connection}

The traditional interpolation algorithm constructs the target function as a linear combination of a series of basis functions, usually each of which corresponds to an observed point. For example, to interpolate $n$ observed points $O=\{(x_i,y_i)\}_{i=1}^n$,
the RBF approach uses the following interpolation function 
\begin{align}
    f_{\mathrm{RBF}}(x)= \sum_{j=1}^n\lambda_j\phi(x,x_j).  \label{eq_rbf}
\end{align}
Here,  $\phi(\cdot,x_j)$ represents a radial basis function specified by the observed point $x_j$, and  $\lambda_j$ is the coefficient, which can be determined through solving the linear equations $\sum_{j=1}^n\lambda_j\phi(x_i,x_j)=y_i$, $(x_i,y_i)\in O$ \cite{solomon2015numerical}. 

NIERT can be cast as such an instance of interpolation algorithm using basis functions by drawing the following parallels. The core of the NIERT interpolator, i.e., the partial self-attention mechanism shown in Eq. (\ref{eq_pattn_layer}) can be rewritten as:
\begin{align}
    f_{\mathrm{Attn}}(x) = \sum_{j=1}^n\alpha(q(x),k_j)v_j.   \label{eq_pattn}
\end{align}
Here, $\alpha(q(x),k_j)$ is the normalized attention weight that models the contribution to query vector $q(x)$ by 
the key vector $k_j$, in which $k_j$ and $v_j$ are dominated by the observed point $x_j$.

Comparing Eq. (\ref{eq_rbf}) with Eq. (\ref{eq_pattn}), we can find that partial self-attention is a general form of RBF interpolation function: $\lambda_j$ in Eq. (\ref{eq_rbf}) corresponds to $v_j$ in Eq. (\ref{eq_pattn}), and $\phi(x,x_j)$ in Eq. (\ref{eq_rbf}) corresponds to $\alpha(q(x),k_j)$ in Eq. (\ref{eq_pattn}).
This correspondence indicates that  $\alpha(q(x),k_j)$ can be treated as a learnable basis function and, similarly, $v_j$ can be treated as a predictable basis function coefficient. From this point of view, partial self-attention is a learnable layer that interpolates the representation of observed points to yield new representations of data points. Together, this insight provides a plausible explanation of the partial self-attention mechanism, which is mutually supportive with the interpretability analysis of the contribution of observed points shown in Section \ref{subsec_contrib}.

\section{Conclusion}

\label{sec:conclusion}

We present in the study an accurate approach to numerical interpolation for scattered data. The specific features of our NIERT approach are highlighted by the full exploitation of the correlation between observed points and target points through unifying scattered data representation. At the same time, the use of the partial self-attention mechanism can effectively avoid the interference of target points onto the observed points. 
We also propose to use the large-scale synthesis of generic and diverse mathematical functions to build pre-trained NIERT interpolators with general interpolation capabilities.
The advantages of NIERT in interpolation accuracy have been clearly demonstrated by experimental results on both synthetic and real-world datasets. 
We expect NIERT, with extensions and modifications, to greatly facilitate numerical interpolations in a wide range of engineering and science fields. 

\ifCLASSOPTIONcompsoc
  \section*{Acknowledgments}
\else
  \section*{Acknowledgment}
\fi

We would like to thank the National Key Research and Development Program of China (2020YFA0907000), the National Natural Science Foundation of China (32271297, 62072435, 82130055), and the Leading Innovative and Entrepreneur Team Introduction Program of Zhejiang (2019R02002) for providing financial supports for this study and publication charges.

\ifCLASSOPTIONcaptionsoff
  \newpage
\fi



%

\bibliographystyle{IEEEtran}
\bibliography{content/ref}

%

\begin{IEEEbiography}[{\includegraphics[width=1in,height=1.25in,clip,keepaspectratio]{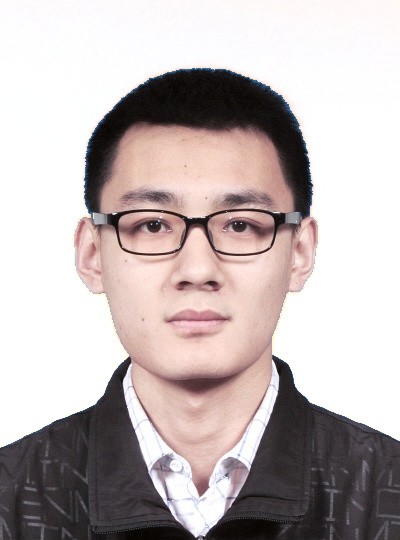}}]{Shizhe Ding}
is currently pursuing his Ph.D. degree in the Institute of Computing Technology (ICT), Chinese Academy of Sciences (CAS), Beijing, China. He recieved his B.E. degree from University of Chinese Academy of Sciences in 2019. His research interests lie in data mining and deep Learning.
\end{IEEEbiography}

\begin{IEEEbiography}[{\includegraphics[width=1in,height=1.25in,clip,keepaspectratio]{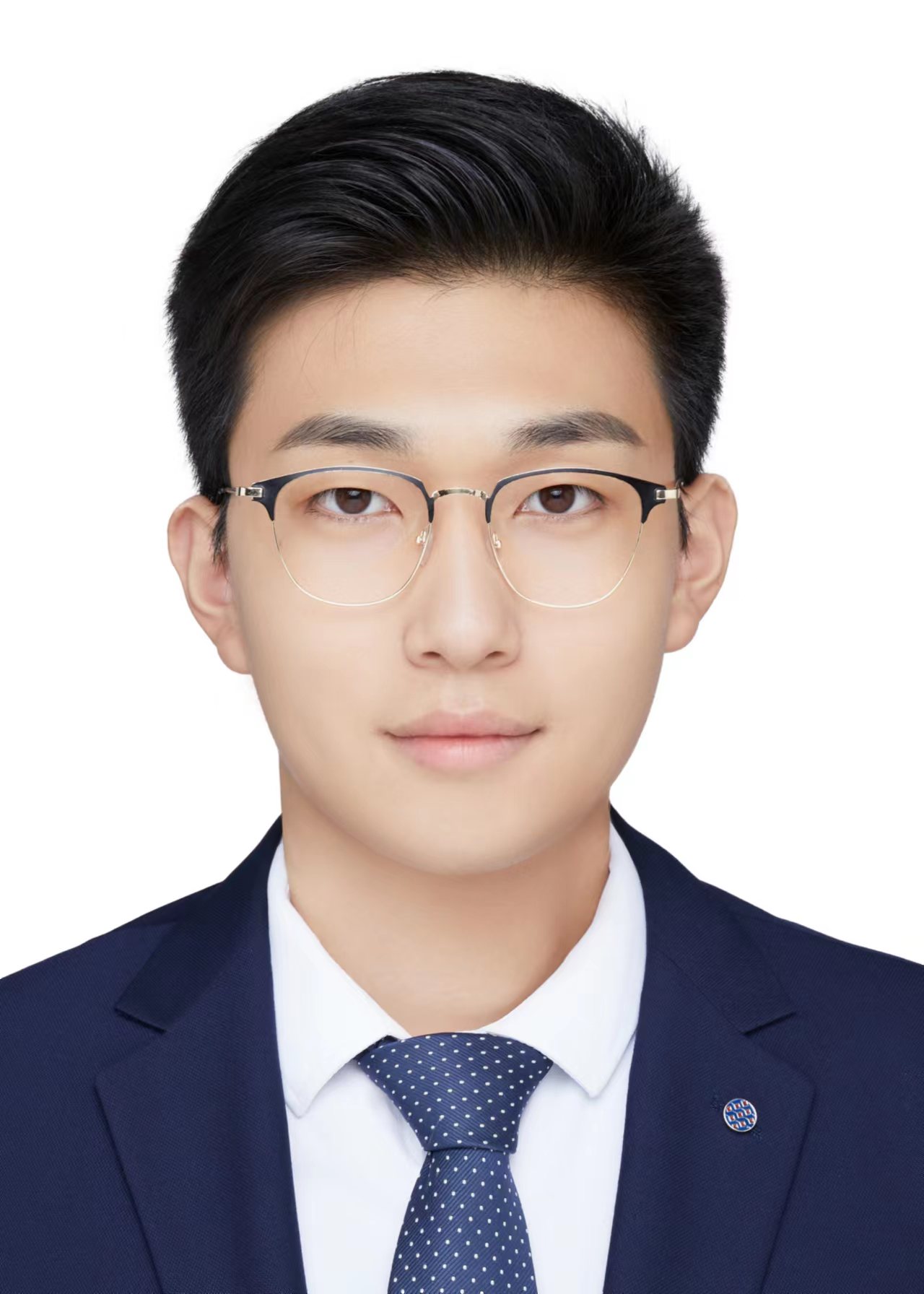}}]{Boyang Xia}
 received B.S. degree in 2020 from Beijing University of Posts and Telecommunications. He is currently a master candidate of the Institute of Computing Technology (ICT), Chinese Academy of Sciences (CAS). His research interests include deep learning and computer vision.
\end{IEEEbiography}

\begin{IEEEbiography}[{\includegraphics[width=1in,height=1.25in,clip,keepaspectratio]{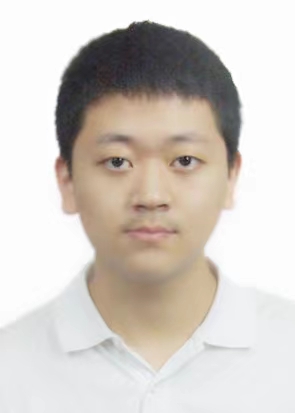}}]{Milong Ren}
received B.S. degree from Shandong University in 2021. He is currently pursuing his Ph.D. degree in the Institute of Computing Technology (ICT), Chinese Academy of Sciences (CAS). His research interests include deep learning and bioinfomatics.
\end{IEEEbiography}

\begin{IEEEbiography}[{\includegraphics[width=1in,height=1.25in,clip,keepaspectratio]{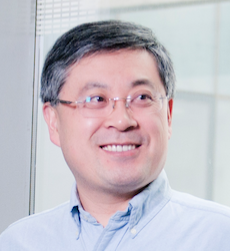}}]{Dongbo Bu} recieved his Ph. D. degree from the Institute of Computing Technology (ICT), Chinese Academy of Sciences (CAS). He is now a professor at the institute and his research interests include AI-aided algorithm design, algorithms for protein structure prediction and protein design. He worked with his colleagues to develop ProFOLD, a software for protein structure prediction, ProDESIGN, a software for protein design, and AIA-ILP, a software for solving integer linear programming with aid of AI.
\end{IEEEbiography}


\vfill


\newpage
\ 
\newpage

\onecolumn
\section{Additional experimental results}

\subsection{Additional ablation studies}

\subsubsection{The effects of different model depths}

We carried out experiments on Mathit-2D dataset using NIERT with hyper-parameter $L$ varying from $3$ to $7$. Then evaluate the models on the Mathit-2D test dataset. The accuracy is listed in Table~\ref{table:ablation_layers}. The results show that NIERT with $7$ transformer layers has the best accuracy on the test set, and NIERT with $6$ transformer layers has reached a comparable level. Therefore, in the experiments on Mathit data set, we use $L = 6$ to balance efficiency and accuracy.

\begin{table}[hbt!]\centering
    \caption{The interpolation accuracy of NIERT (MSE $\times 10^{-5}$) under various settings of  transformer layer number $L$  on Mathit-2D dataset}
    \begin{tabular}{crrrrr} \toprule
        \multirow{2}{*} [-2pt] { \makecell{Interpolation\\approach} } &   \multicolumn{5}{c}{ Number of transformer layers $L$} \\ \cmidrule{2-6}
        & \multicolumn{1}{c}{3} & \multicolumn{1}{c}{4} & \multicolumn{1}{c}{5} & \multicolumn{1}{c}{6} & \multicolumn{1}{c}{7} \\  \midrule
        NIERT     & 66.812  & 60.133  & 52.098  & 45.319  & \textbf{44.043} \\ \bottomrule
    \end{tabular}
    \label{table:ablation_layers}
\end{table}

\subsubsection{The effects of different hidden dimensions}

We also carried out experiments on the Mathit-2D dataset using NIERT with smaller hidden dimensions $d_{model}$, say from $256$, $128$ and $64$. Then evaluate the models on the Mathit-2D test dataset. The accuracy is listed in Table~\ref{table:ablation_widths}. The results show that when the hidden dimension is within 512, NIERT's interpolation accuracy is higher when the hidden dimension is larger.

In this experiment, we fix other super parameters, say model depth $L$ as $6$ and number of heads as $8$.

\begin{table}[hbt!]\centering
    \caption{The interpolation accuracy of NIERT (MSE $\times 10^{-5}$) under various settings of  hidden dimension $d_{model}$  on Mathit-2D dataset}
    \label{table:ablation_widths}
    \begin{tabular}{crrrr} \toprule
        \multirow{2}{*} [-2pt] { \makecell{Interpolation\\approach} } &   \multicolumn{4}{c}{ Different hidden dimensions $d_{model}$} \\ \cmidrule{2-5}
        & \multicolumn{1}{c}{64} & \multicolumn{1}{c}{128} & \multicolumn{1}{c}{256} & \multicolumn{1}{c}{512}  \\  \midrule
        NIERT     & 106.193 & 72.107 & 51.153  & \textbf{45.319} \\ \bottomrule
    \end{tabular}
\end{table}

\subsubsection{The effects of prediction error of observed points in loss function}

To verify the contribution of re-predicting values of the observed points to the interpolation task, we conducted an experiment that puts the prediction error of observed points in the loss function, i.e. only minimizes the estimation error of target points. The results are shown in Table~\ref{table:ablation_if_supervised} which demonstrates that only minimizing estimation error of target points make NIERT performing poorly. This indicates that re-predicting the value of observed points helps NIERT to predict the value of target points more accurately.

\begin{table}[hbt!]\centering
    \caption{Interpolation accuracy (MSE $\times 10^{-5}$) of NIERT trained with loss only containing the prediction error of the target points}
    \label{table:ablation_if_supervised}
    \begin{tabular}{ccc} \toprule
        Loss contains prediction error of     & Only target points & All points  \\ \midrule
        MSE ($\times 10 ^{-5}$)     & 48.931  & \textbf{45.319}  \\ \bottomrule
    \end{tabular}
\end{table}

\onecolumn
\subsection{Additional case studies}
\label{sup:more_cases}

\textbf{Cases from Mathit-1D \& Mathit-2D test set}

In each following example of 1D interpolation task extracted from Mathit test set, the blue curve represents the ground-truth function while the red curves represent the interpolation functions reported by NIERT and the existing approaches.

\begin{figure}[htbp!]
    \centering
    {{\includegraphics[width=15cm]{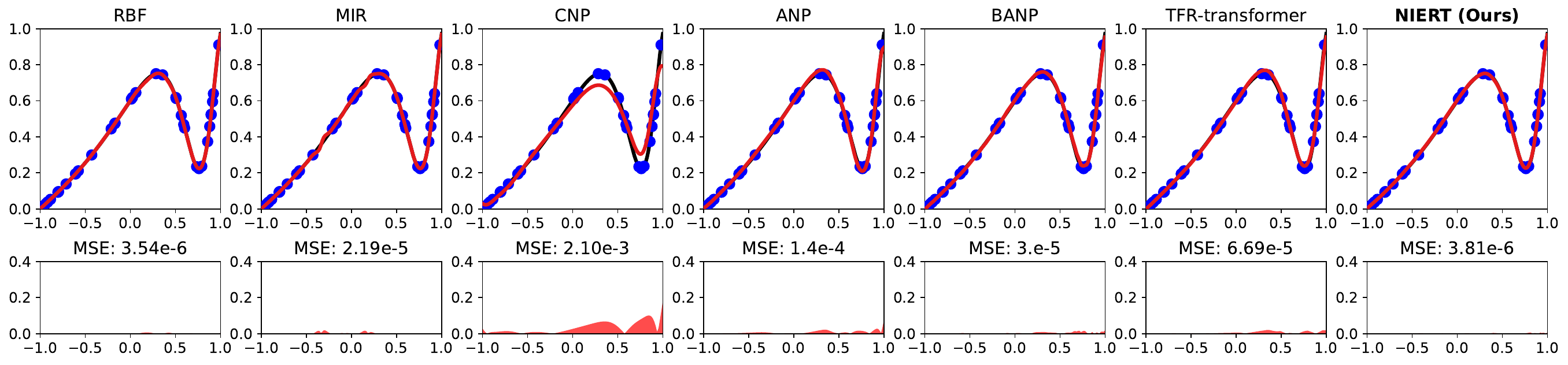} }} \\ \vspace{0.3cm}
    {{\includegraphics[width=15cm]{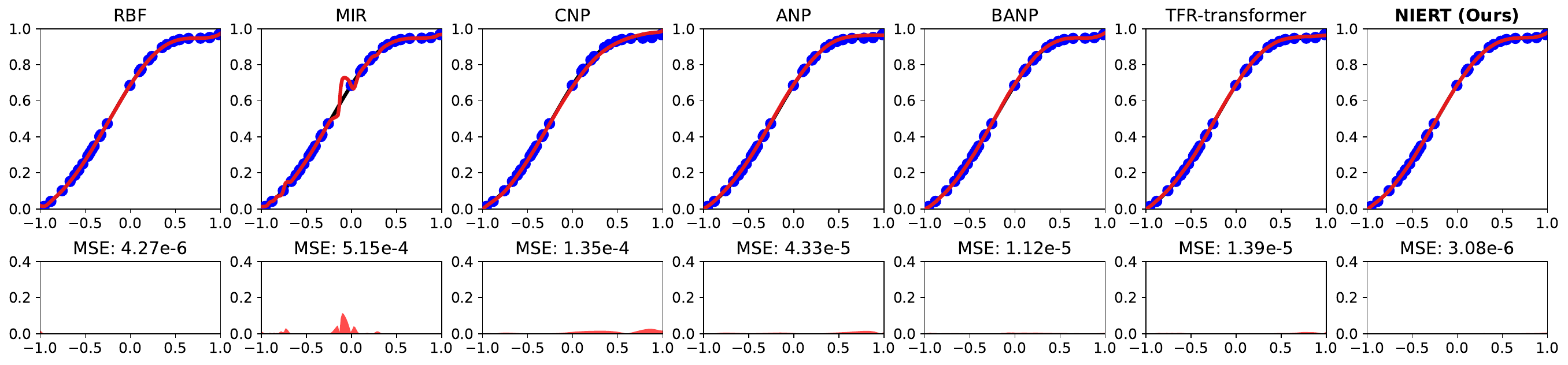} }} \\ \vspace{0.3cm}
    {{\includegraphics[width=15cm]{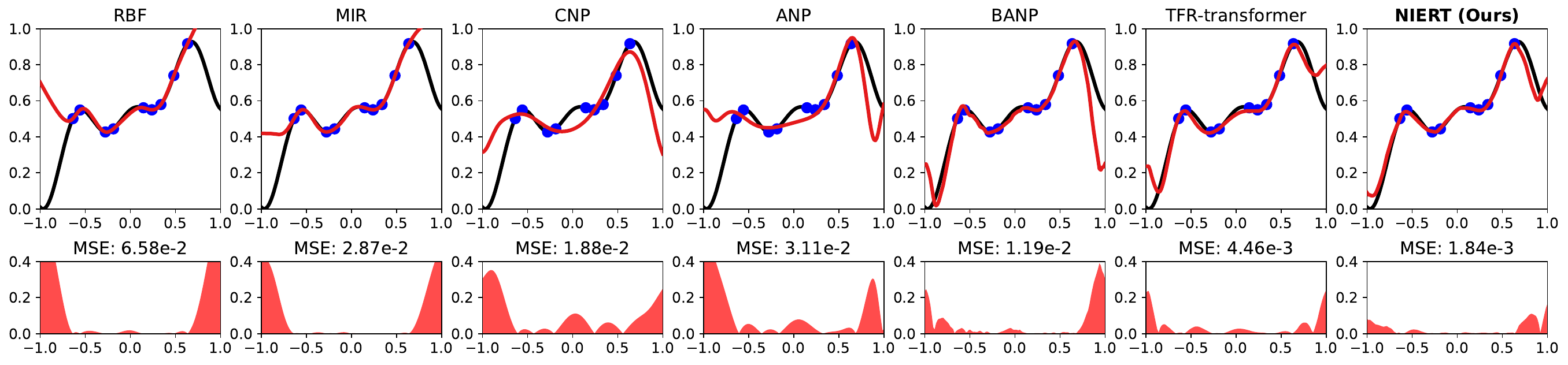} }} \\ \vspace{0.3cm}
    {{\includegraphics[width=15cm]{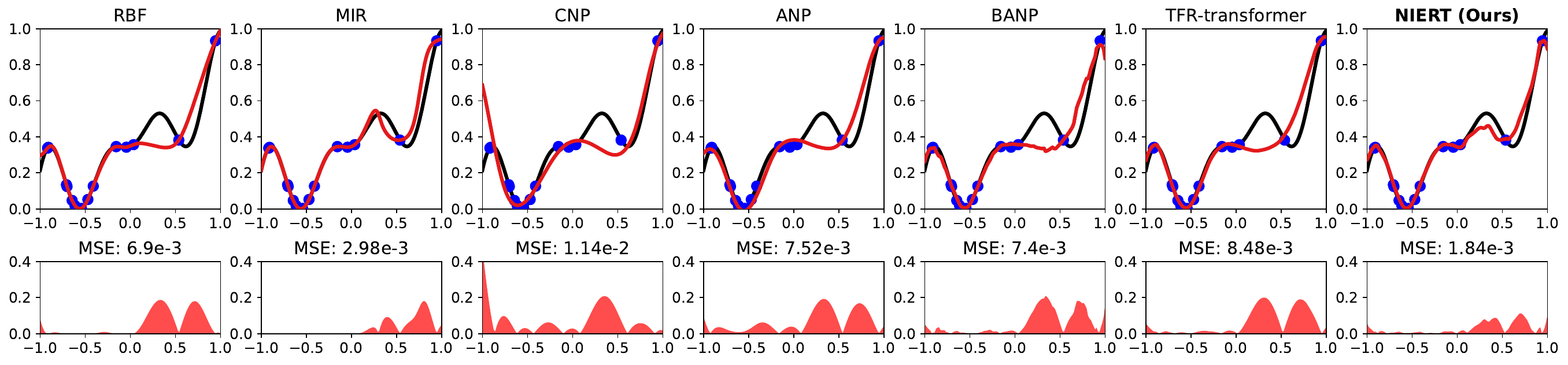} }} \\ \vspace{0.3cm}
    {{\includegraphics[width=15cm]{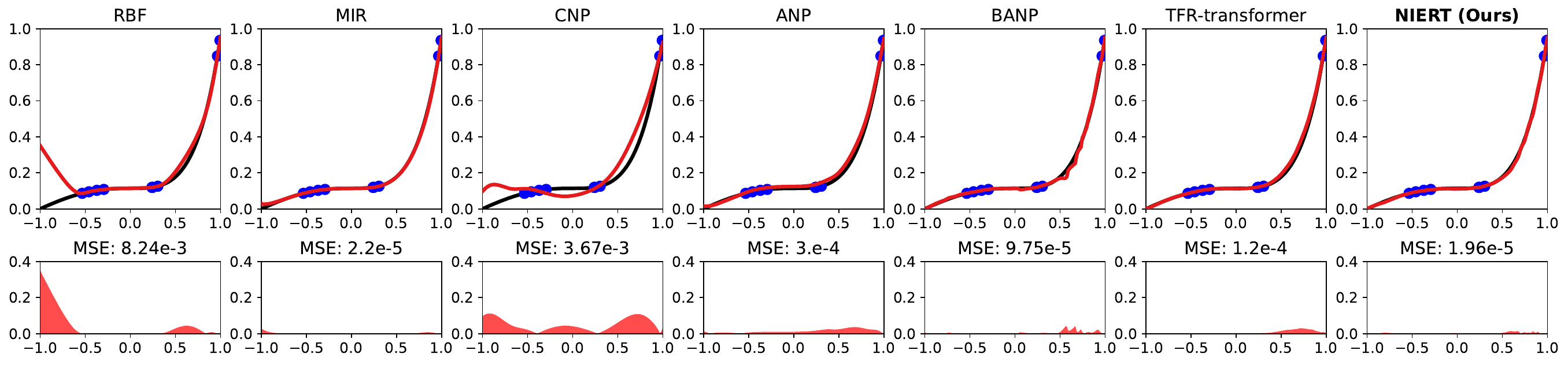} }}
    \caption{Additional cases from MathIT-1D test set}%
    \label{fig:more_1d}%
\end{figure}

\begin{figure}[htbp!]
    \centering
    {{\includegraphics[width=15cm]{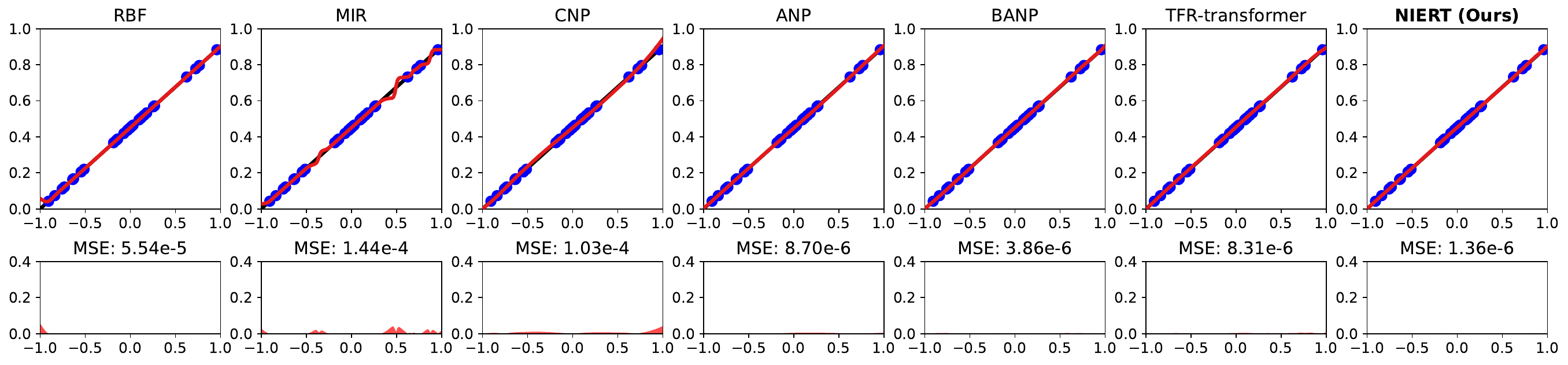} }} \\ \vspace{0.3cm}
    {{\includegraphics[width=15cm]{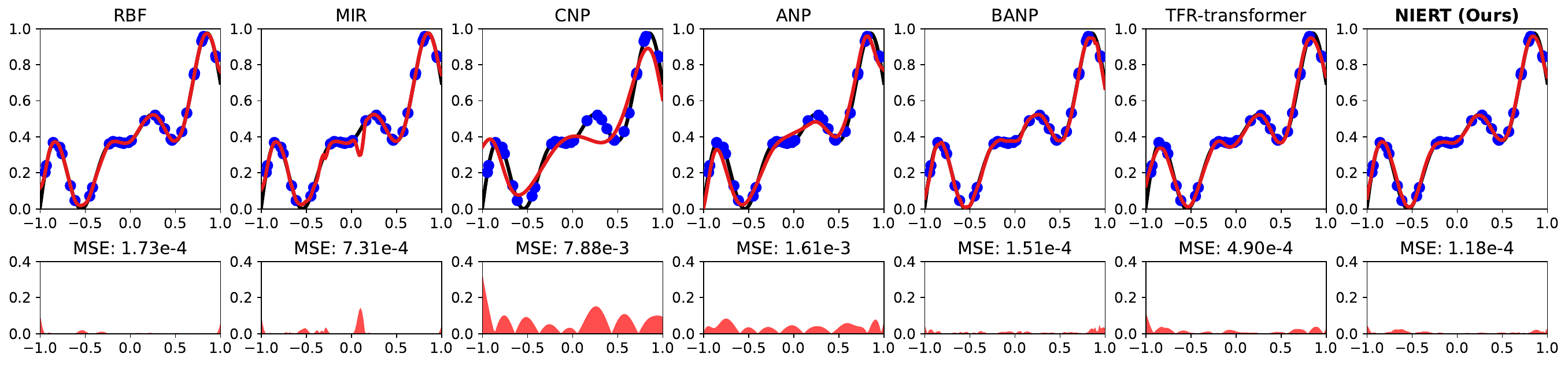} }} \\ \vspace{0.3cm}
    {{\includegraphics[width=15cm]{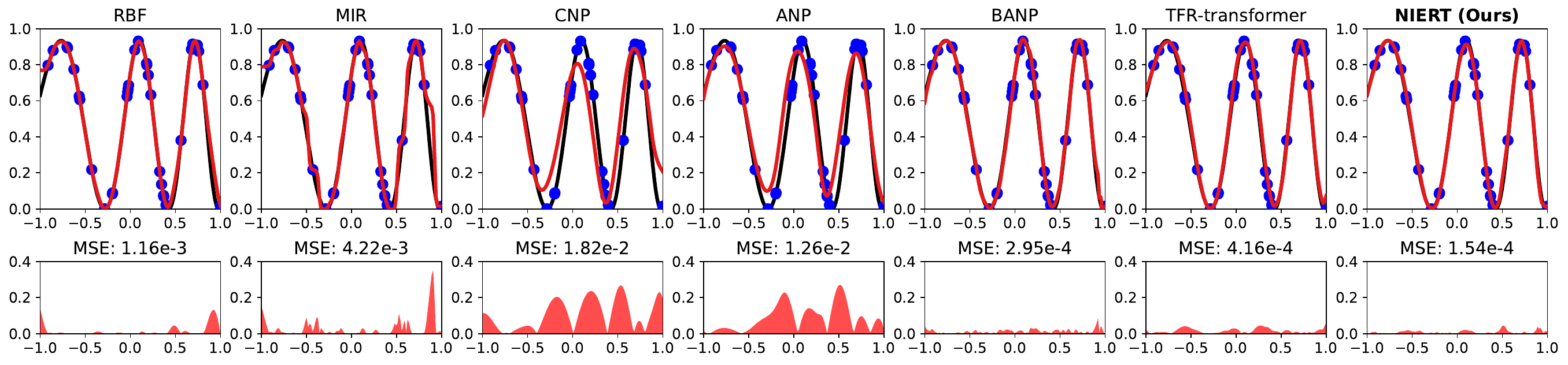} }} \\ \vspace{0.3cm}
    {{\includegraphics[width=15cm]{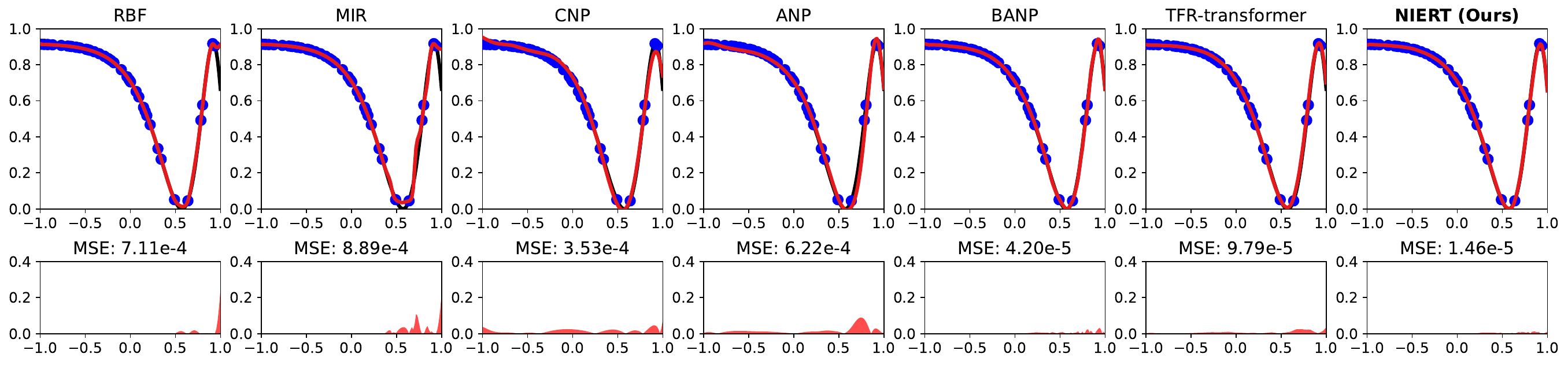} }} \\ \vspace{0.3cm}
    {{\includegraphics[width=15cm]{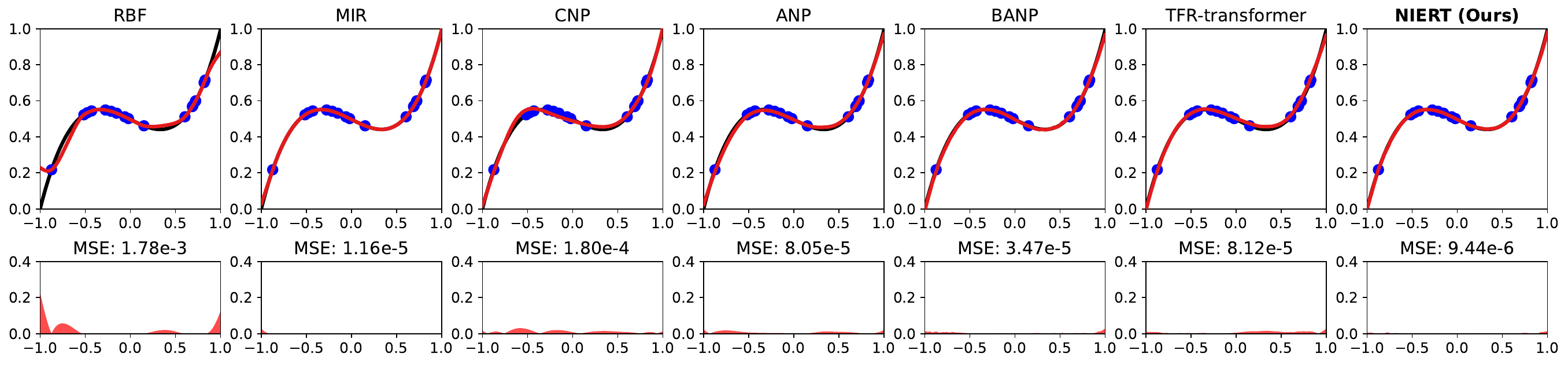} }} 
    \caption{Additional cases from MathIT-1D test set}%
    \label{fig:more_1d_2}%
    \vspace{1cm}
\end{figure}

\begin{figure}[htbp!]
    \centering
    {{\includegraphics[width=15cm]{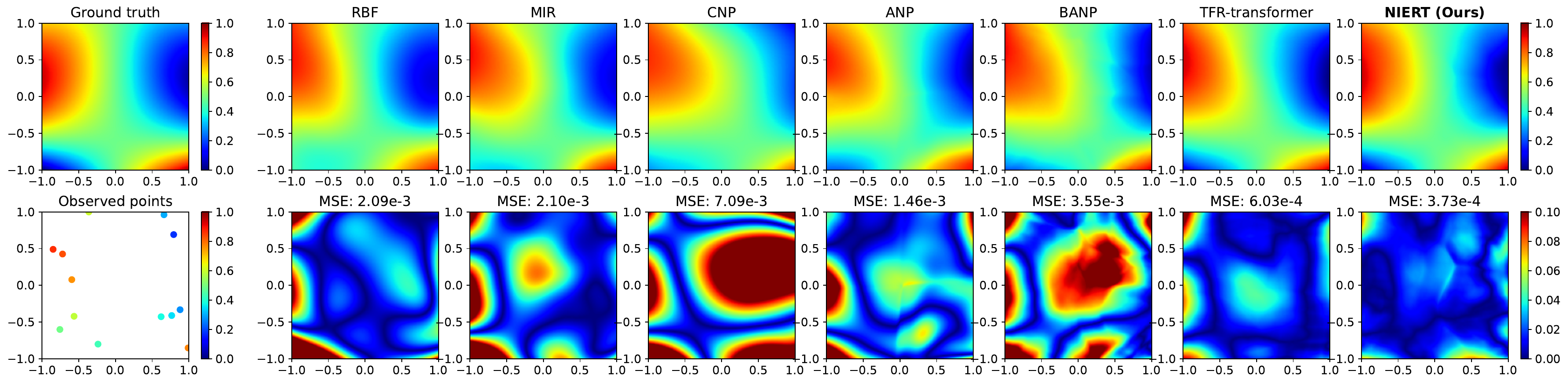} }} \\ \vspace{0.3cm}
    {{\includegraphics[width=15cm]{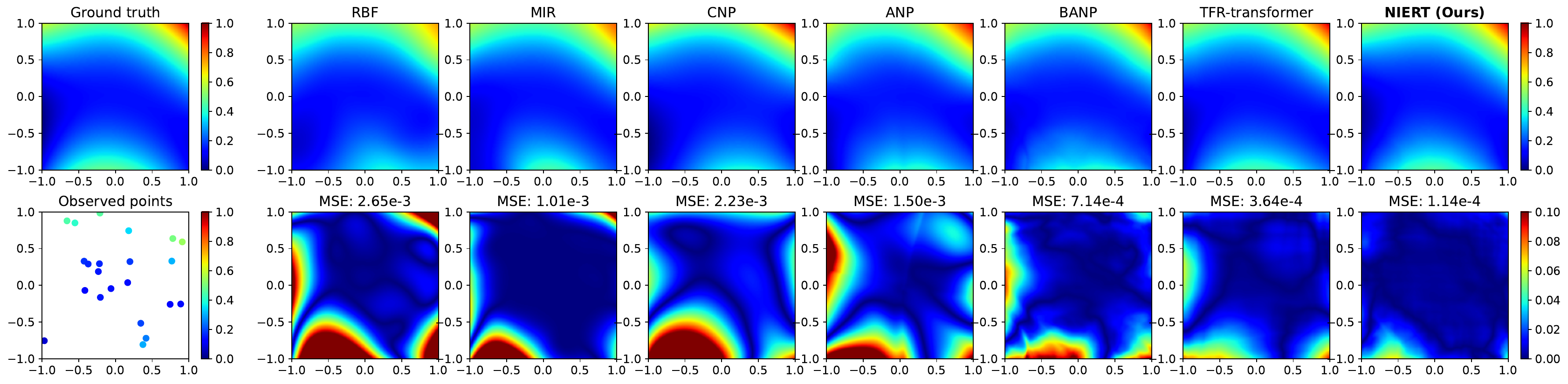} }} \\ \vspace{0.3cm}
    {{\includegraphics[width=15cm]{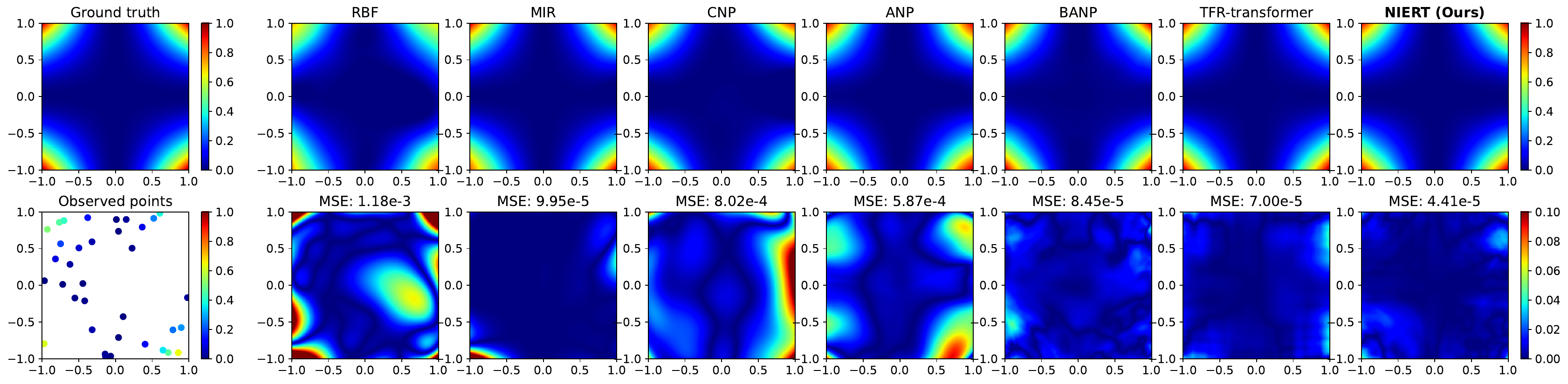} }} \\ \vspace{0.3cm}
    {{\includegraphics[width=15cm]{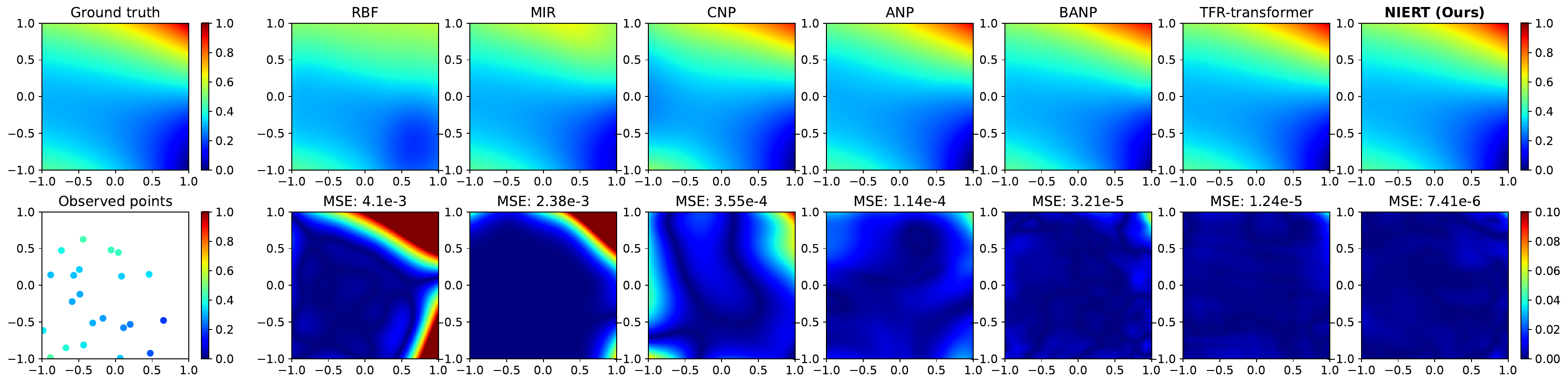} }} \\ \vspace{0.3cm}
    {{\includegraphics[width=15cm]{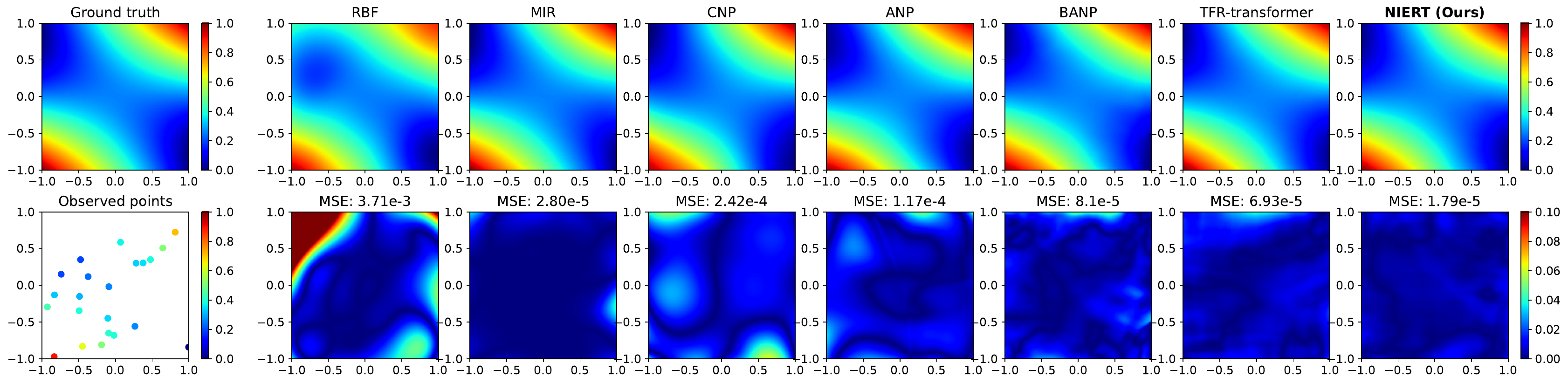} }}
    \caption{Additional cases from MathIT-2D test set}%
    \label{fig:more_2d}%
\end{figure}

\begin{figure}[htbp!]
    \centering
    {{\includegraphics[width=15cm]{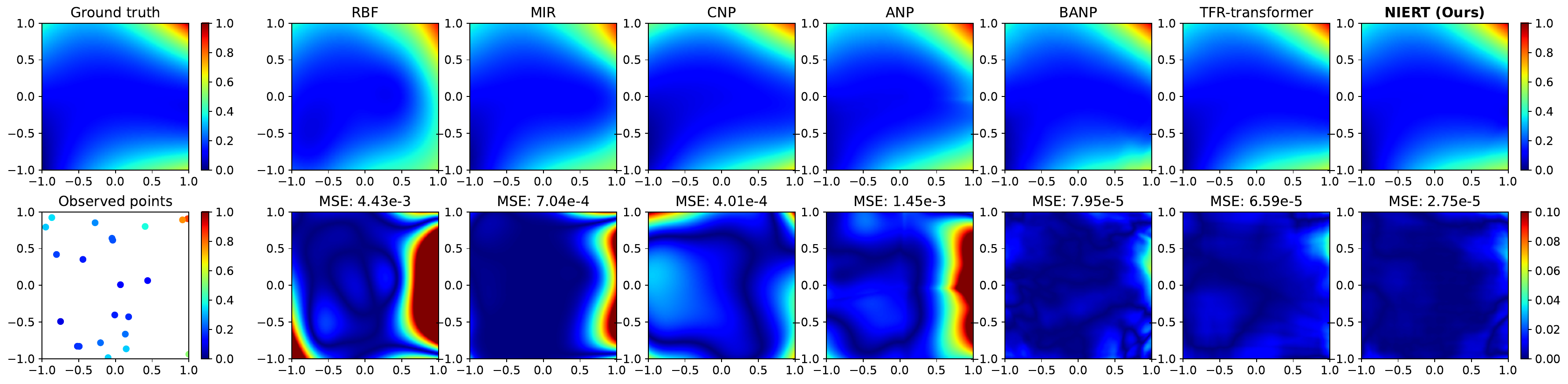} }} \\ \vspace{0.3cm}
    {{\includegraphics[width=15cm]{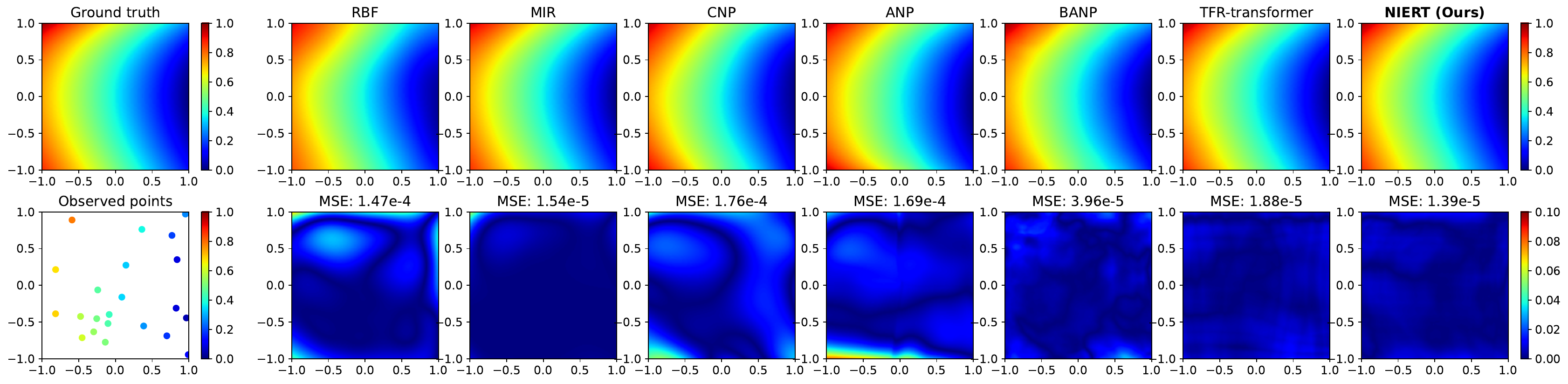} }} \\ \vspace{0.3cm}
    {{\includegraphics[width=15cm]{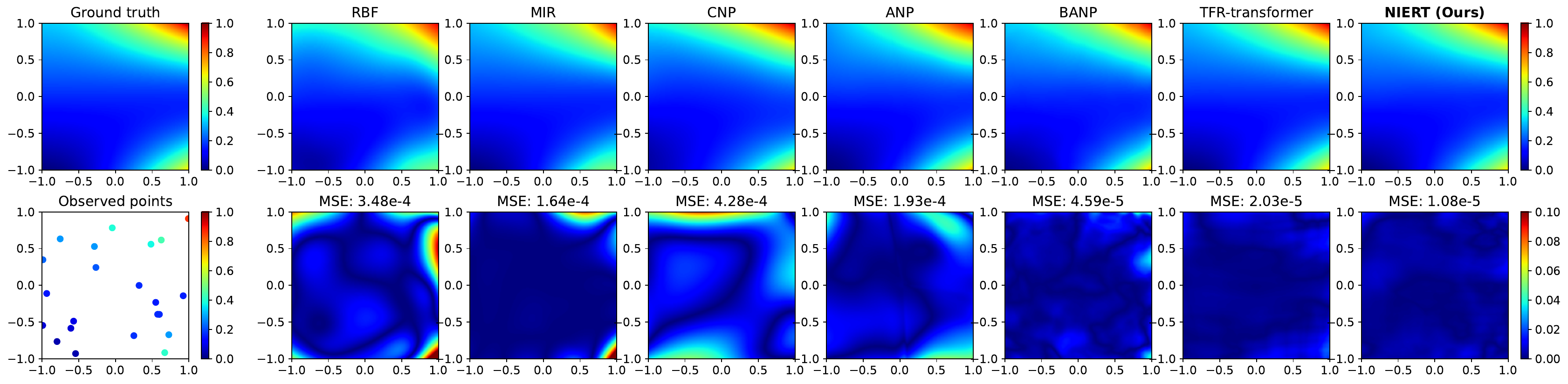} }} \\ \vspace{0.3cm}
    {{\includegraphics[width=15cm]{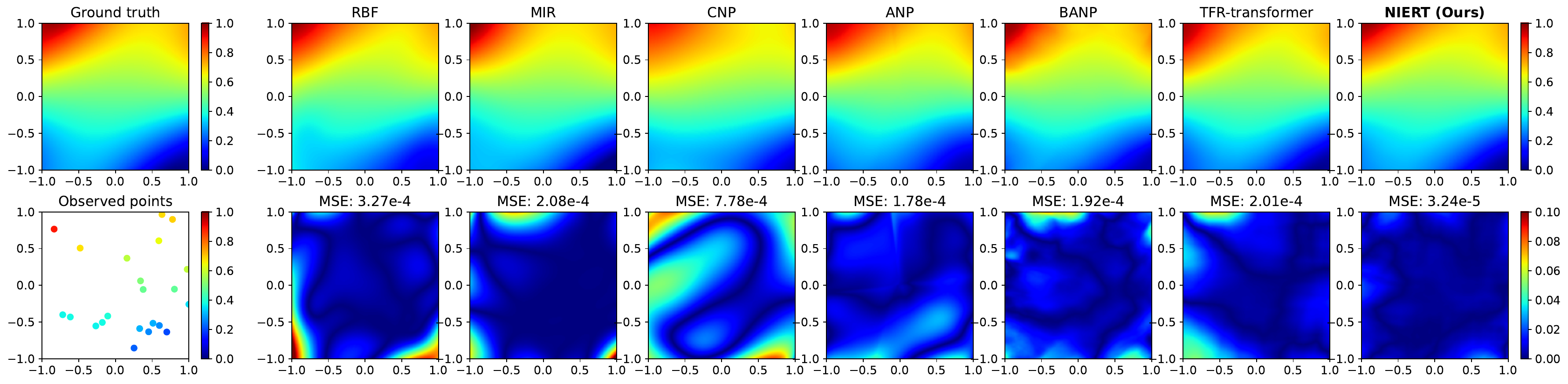} }} \\ \vspace{0.3cm}
    {{\includegraphics[width=15cm]{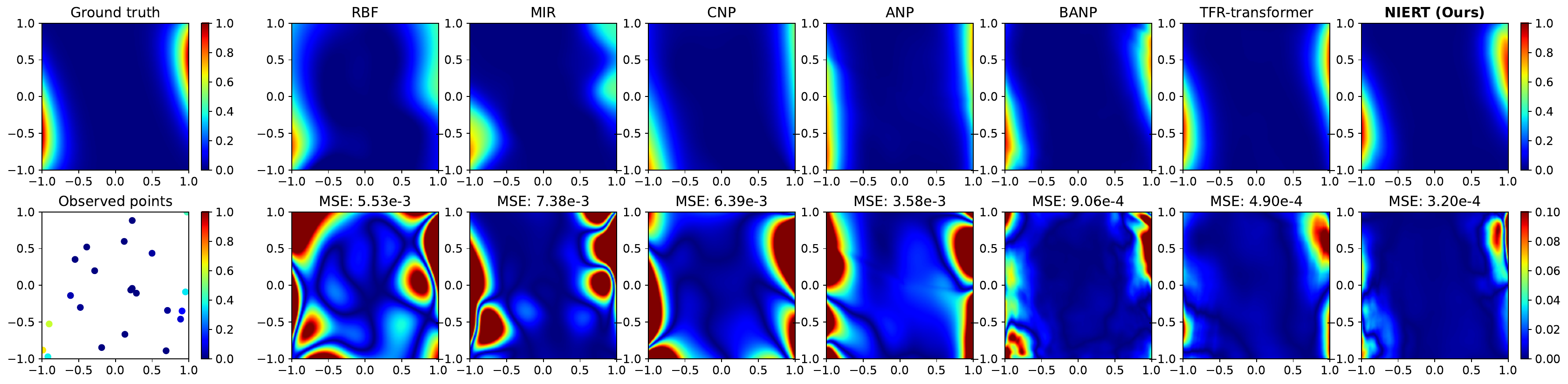} }}
    \caption{Additional cases from MathIT-2D test set}%
    \label{fig:more_2d_2}%
\end{figure}

\newpage
\newpage

\textbf{Cases from TFRD-HSink}

\begin{figure}[htbp!]
    \centering
    {{\includegraphics[width=15cm]{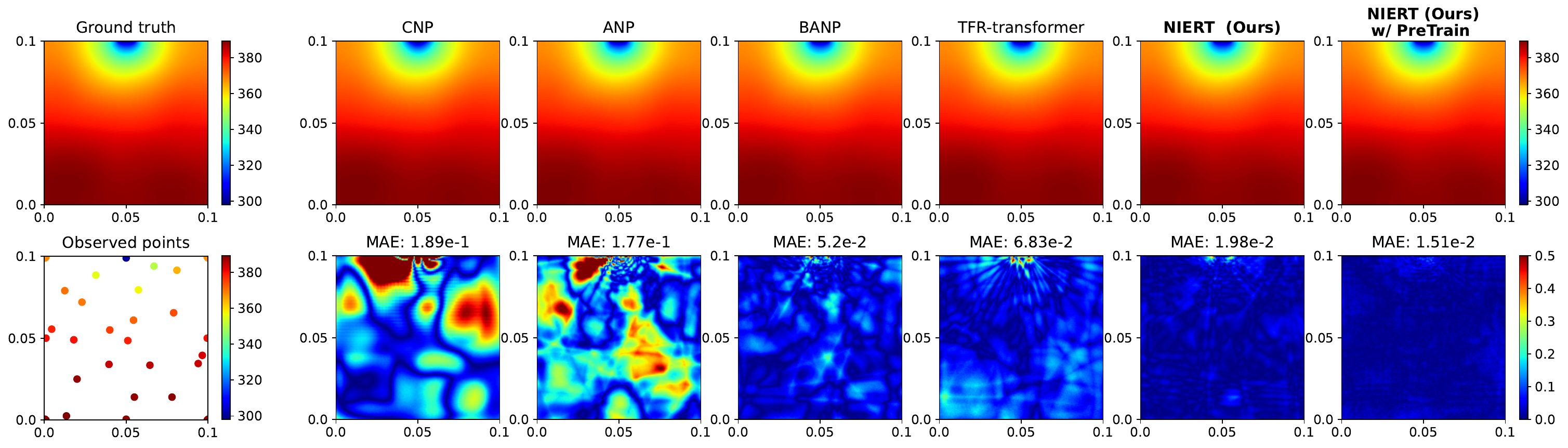} }} \\
    \vspace{0.25cm}
    {{\includegraphics[width=15cm]{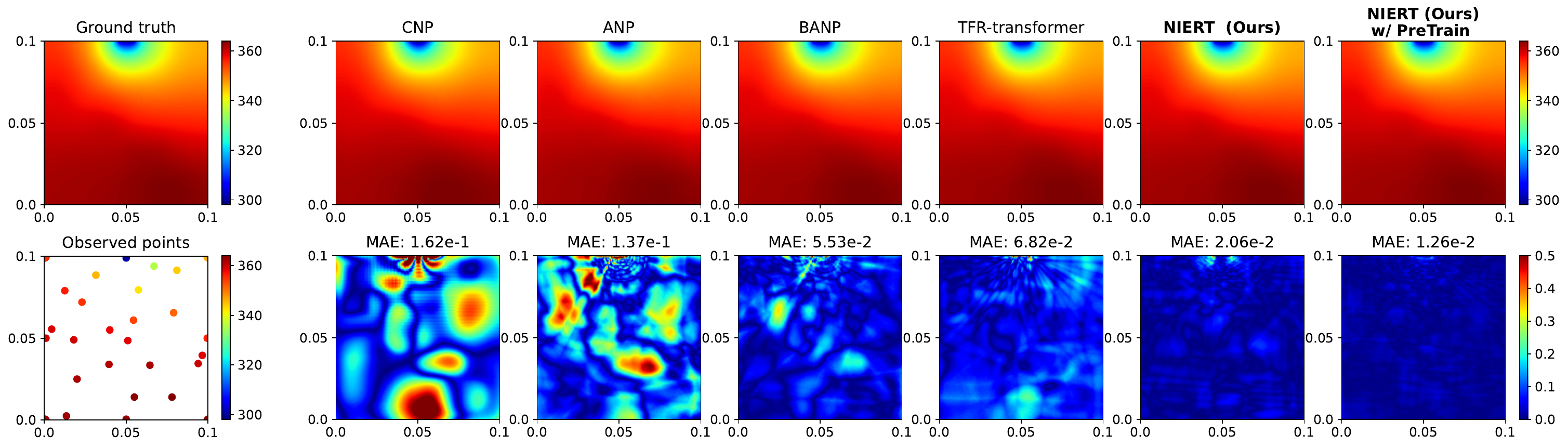} }} \\
    \vspace{0.25cm}
    {{\includegraphics[width=15cm]{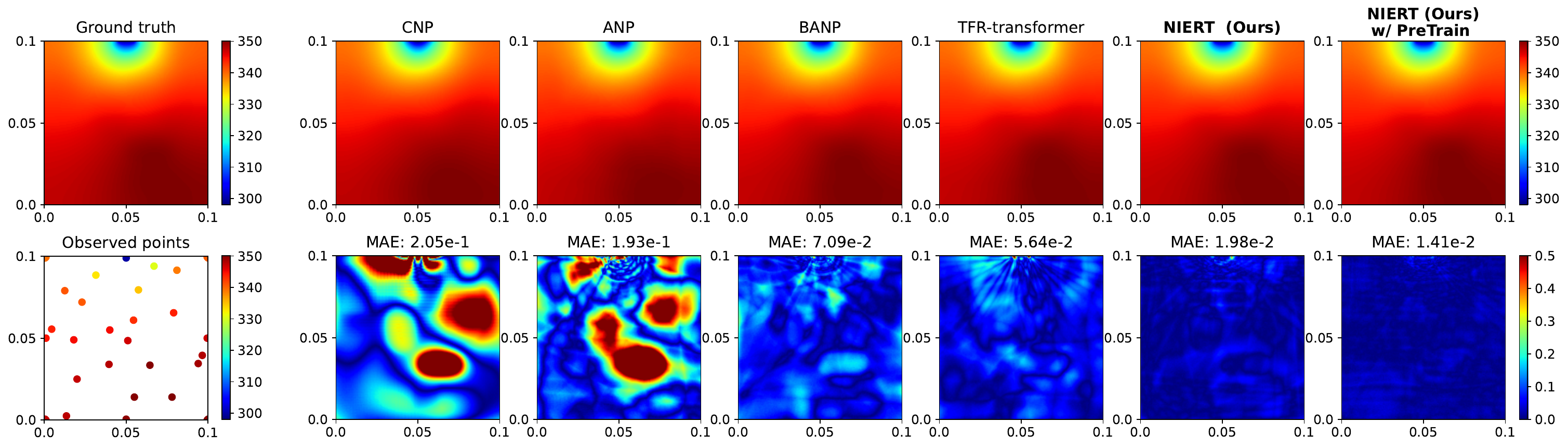} }} \\
    \vspace{0.25cm}
    {{\includegraphics[width=15cm]{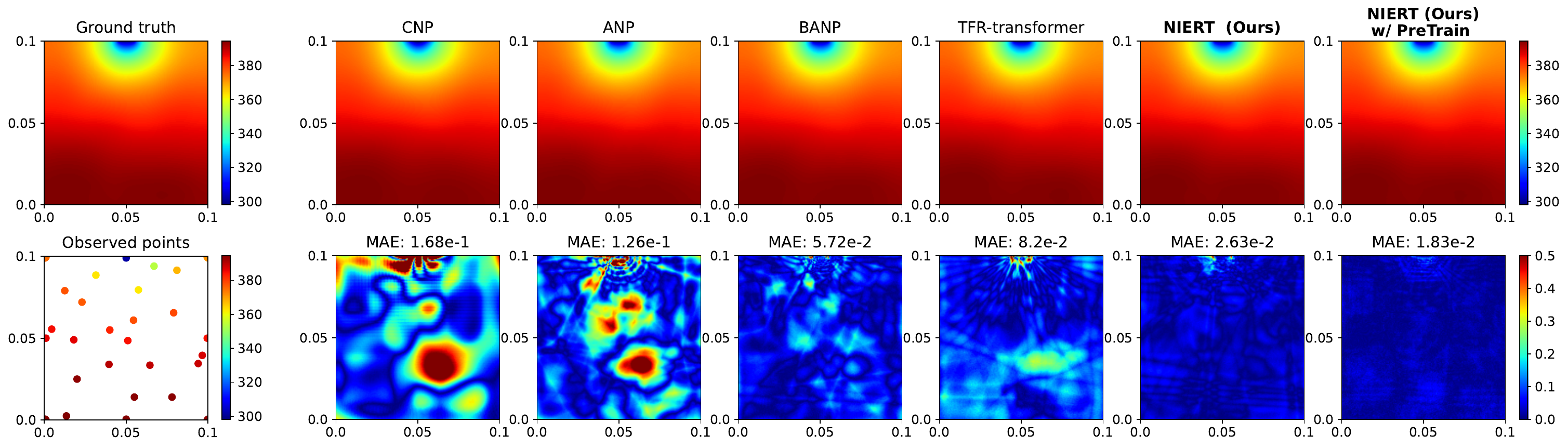} }} \\
    \vspace{0.25cm}
    {{\includegraphics[width=15cm]{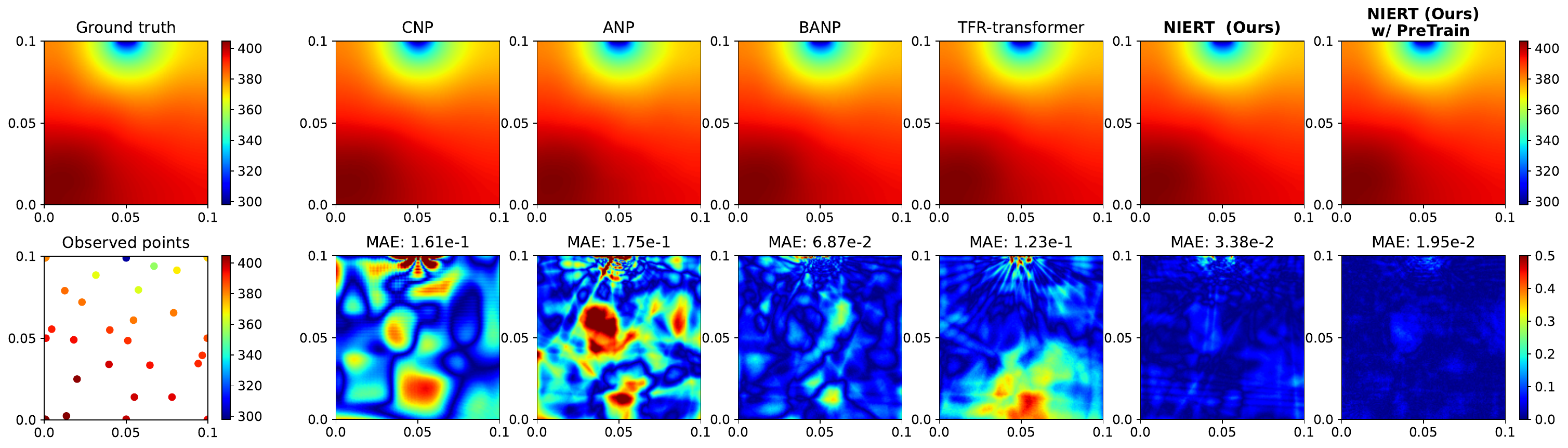} }} \\
    \caption{Additional cases from TFRD-HSink test set}%
    \label{fig:more_tfrd_hsink}%
\end{figure}

\newpage

\textbf{Cases from TFRD-ADlet}

\begin{figure}[htbp!]
    \centering
    {{\includegraphics[width=15cm]{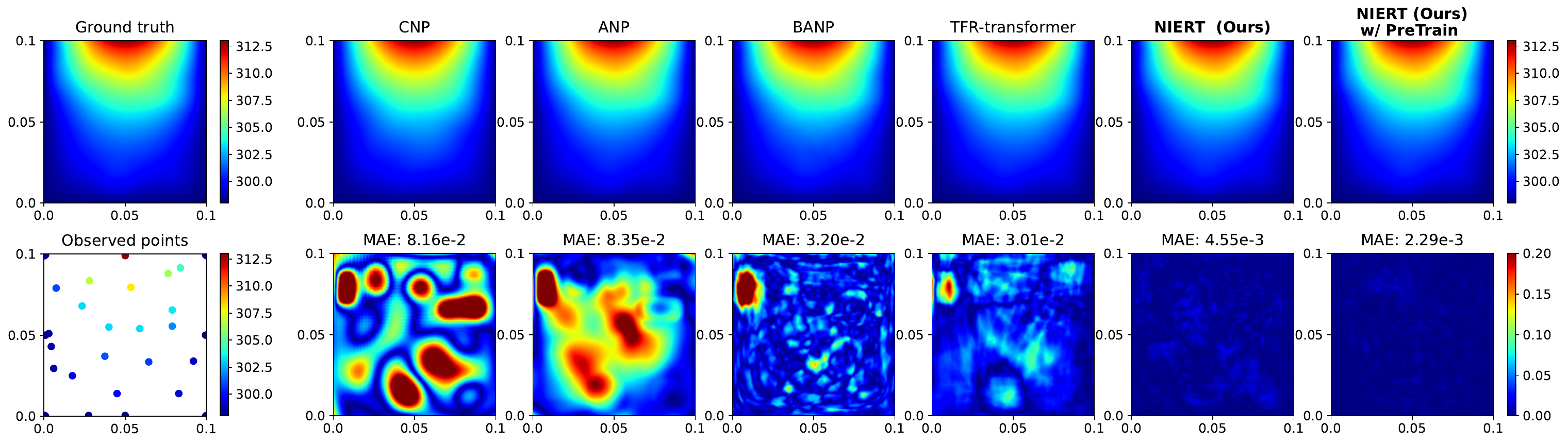} }} \\
    \vspace{0.25cm}
    {{\includegraphics[width=15cm]{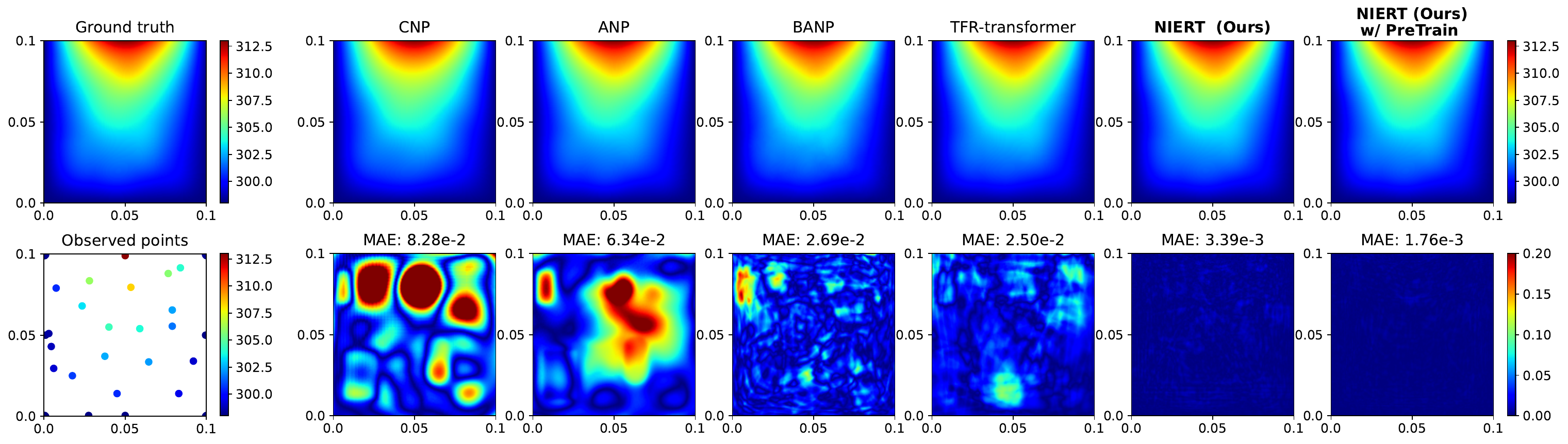} }} \\
    \vspace{0.25cm}
    {{\includegraphics[width=15cm]{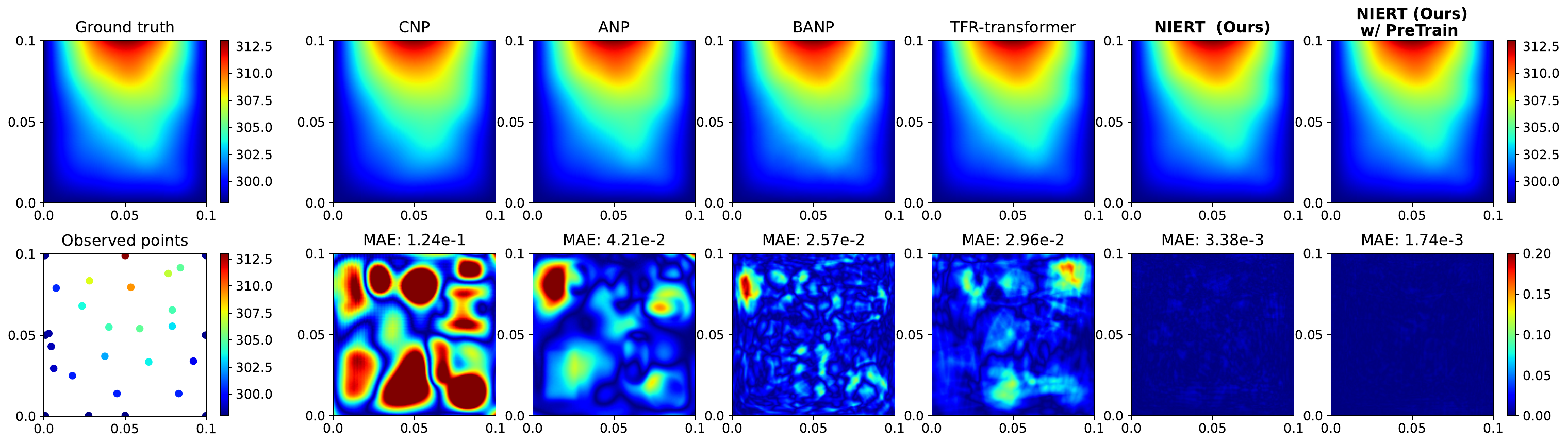} }} \\
    \vspace{0.25cm}
    {{\includegraphics[width=15cm]{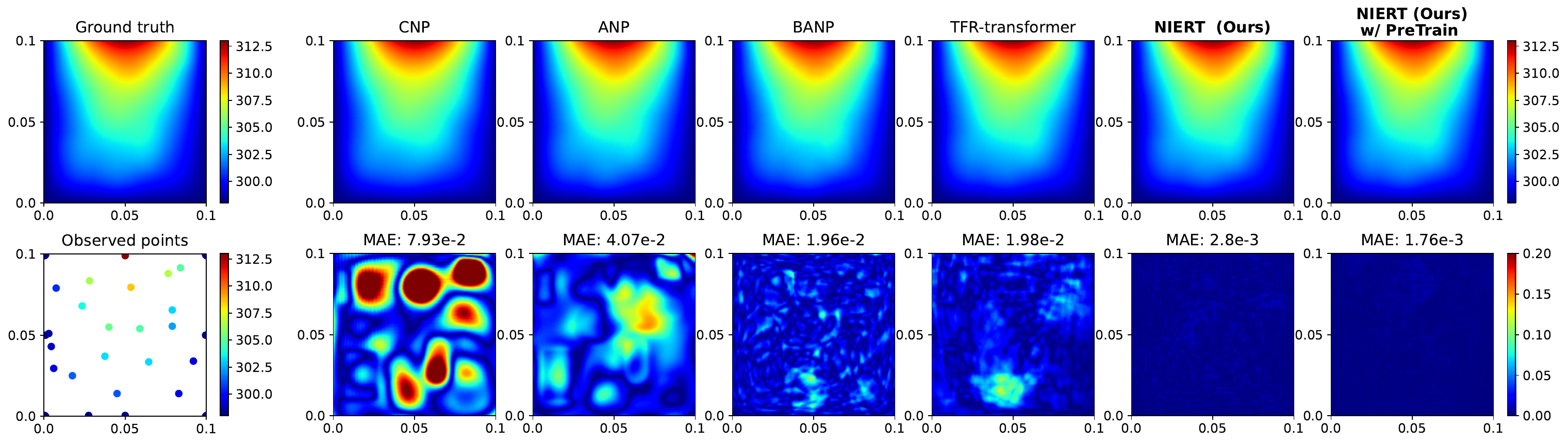} }} \\
    \vspace{0.25cm}
    {{\includegraphics[width=15cm]{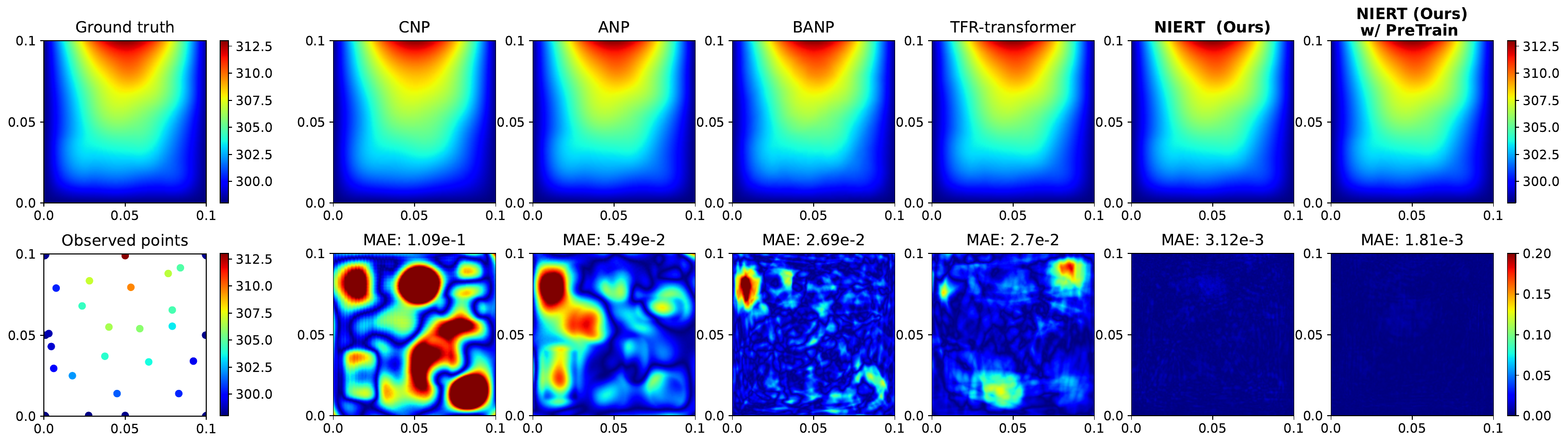} }} \\
    \caption{Additional cases from TFRD-ADlet test set}%
    \label{fig:more_tfrd_adlet}%
\end{figure}

\newpage
\textbf{Cases from TFRD-DSine}

\begin{figure}[htbp!]
    \centering
    {{\includegraphics[width=15cm]{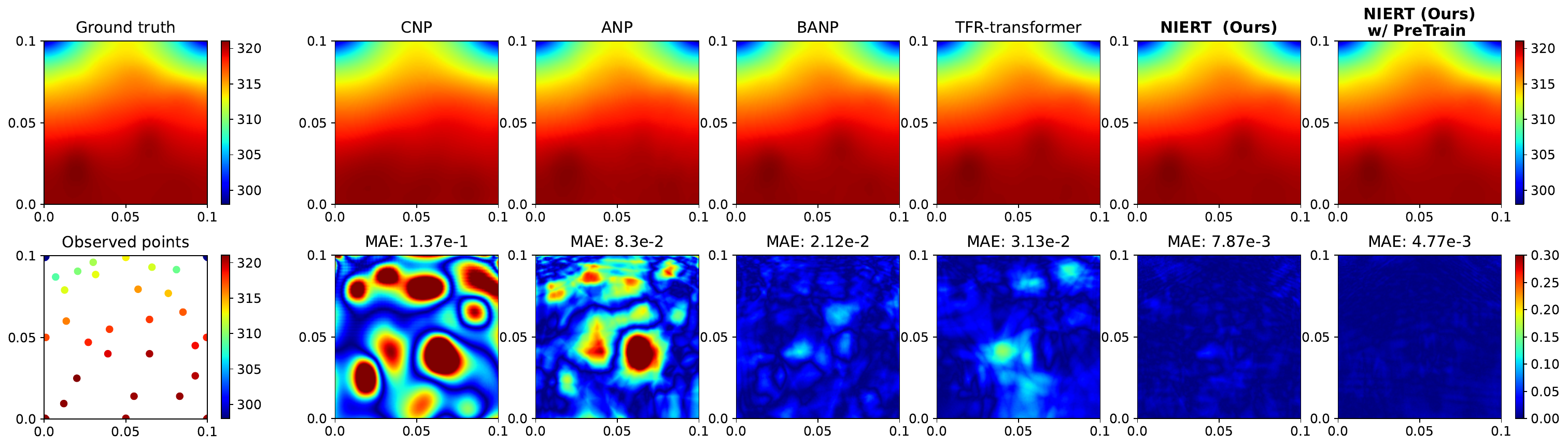} }} \\
    \vspace{0.25cm}
    {{\includegraphics[width=15cm]{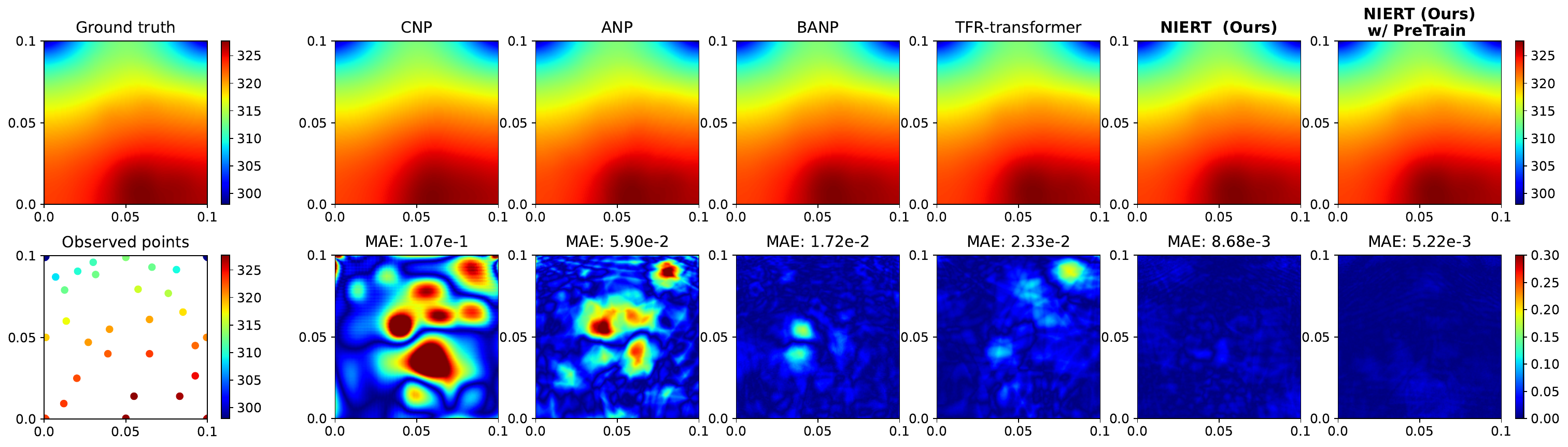} }} \\
    \vspace{0.25cm}
    {{\includegraphics[width=15cm]{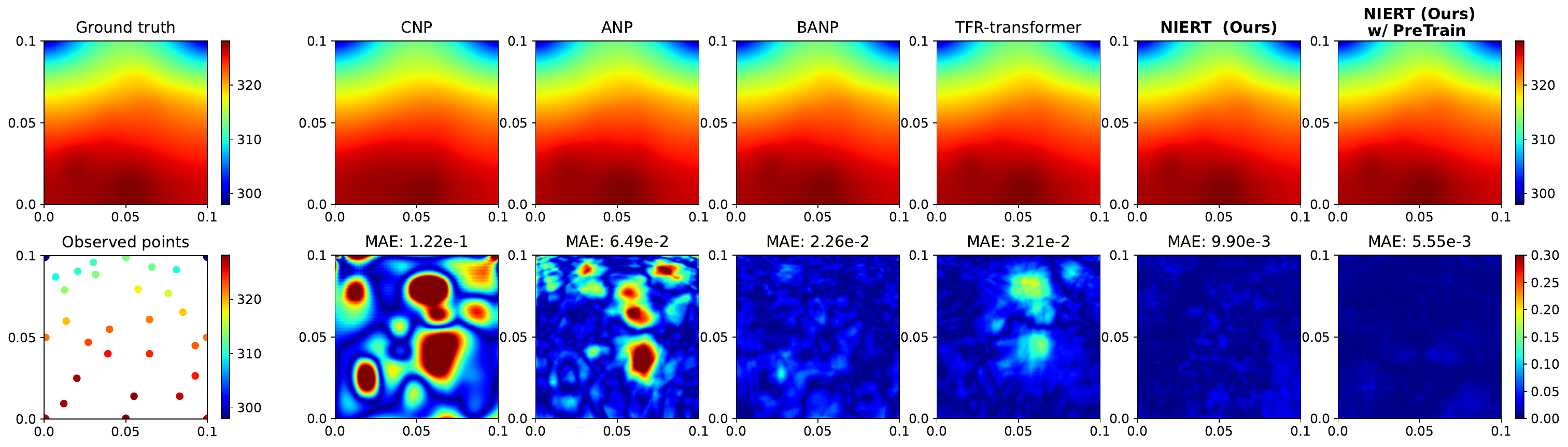} }} \\
    \vspace{0.25cm}
    {{\includegraphics[width=15cm]{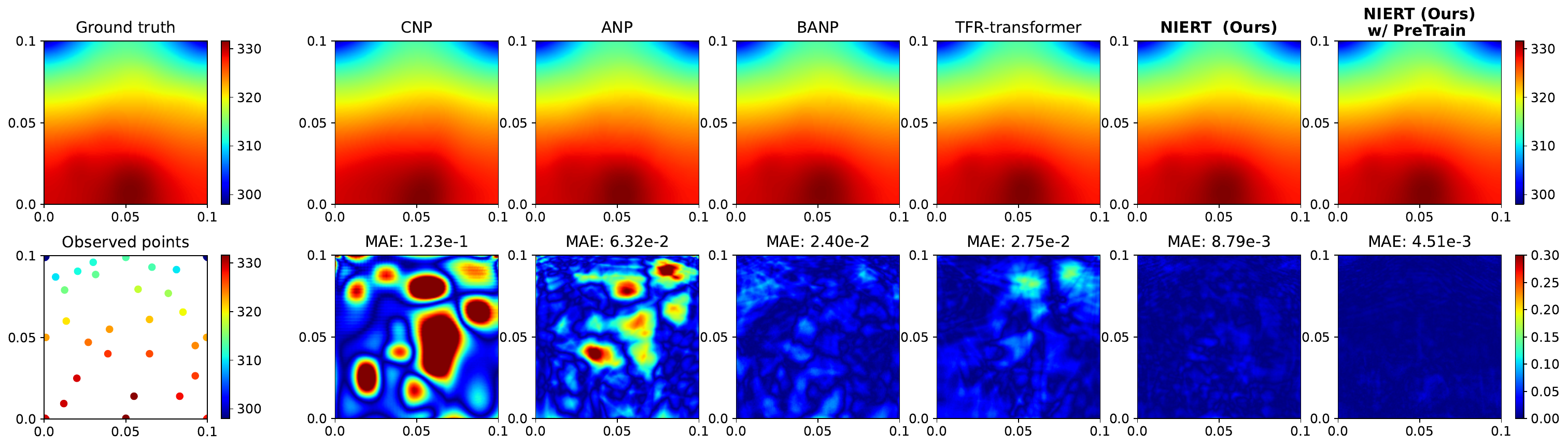} }} \\
    \vspace{0.25cm}
    {{\includegraphics[width=15cm]{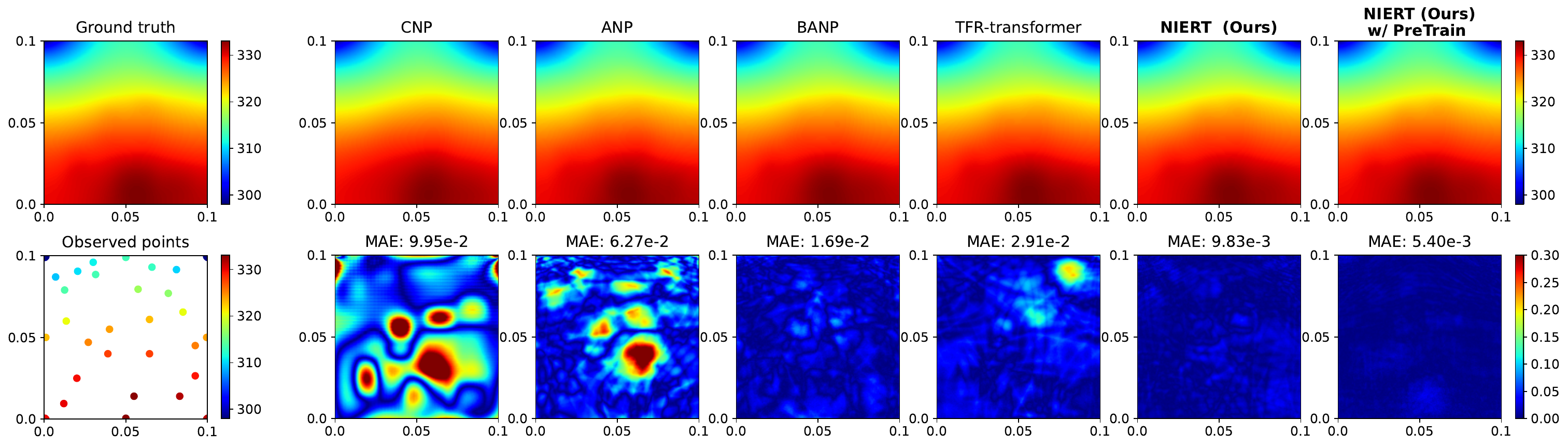} }} \\
PIV    \caption{Additional cases from TFRD-DSine test set}%
    \label{fig:more_tfrd_dsine}%
\end{figure}

\newpage
\textbf{Cases from PTV}

\begin{figure}[htbp!]
    \centering
    {{\includegraphics[width=15cm]{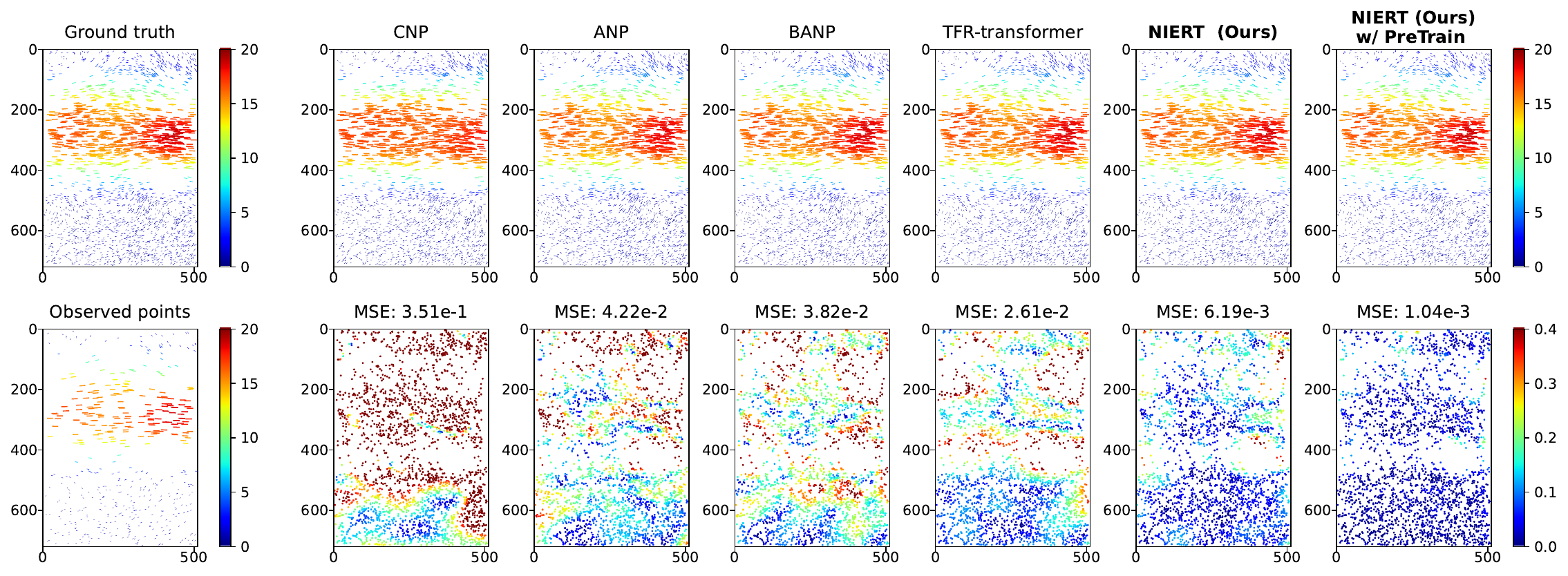} }} \\
    \vspace{0.3cm}
    {{\includegraphics[width=15cm]{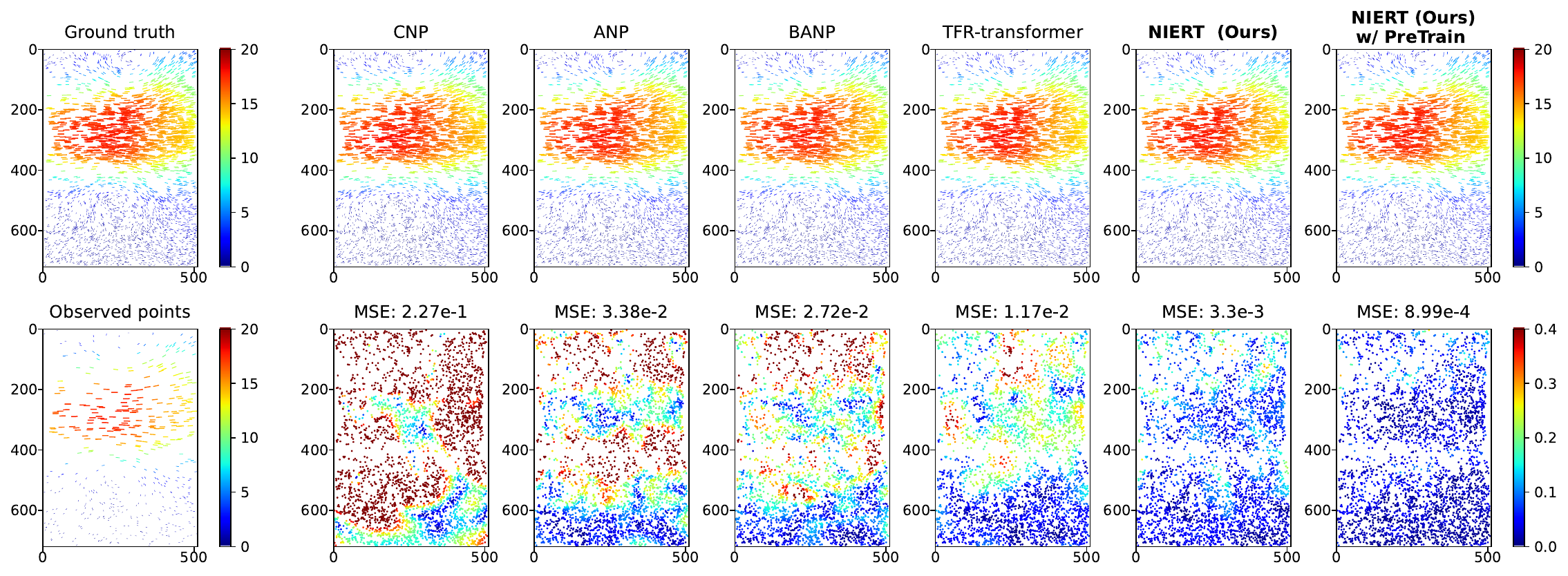} }} \\
    \vspace{0.3cm}
    {{\includegraphics[width=15cm]{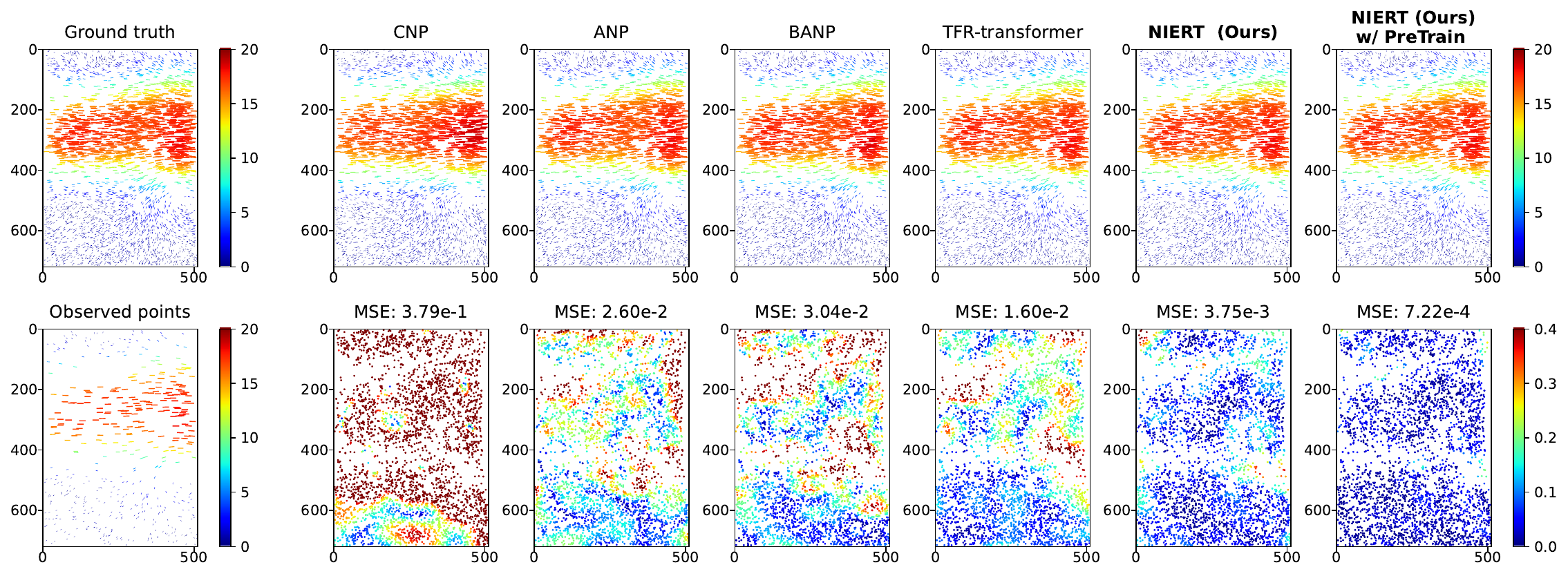} }} \\
    \vspace{0.3cm}
    {{\includegraphics[width=15cm]{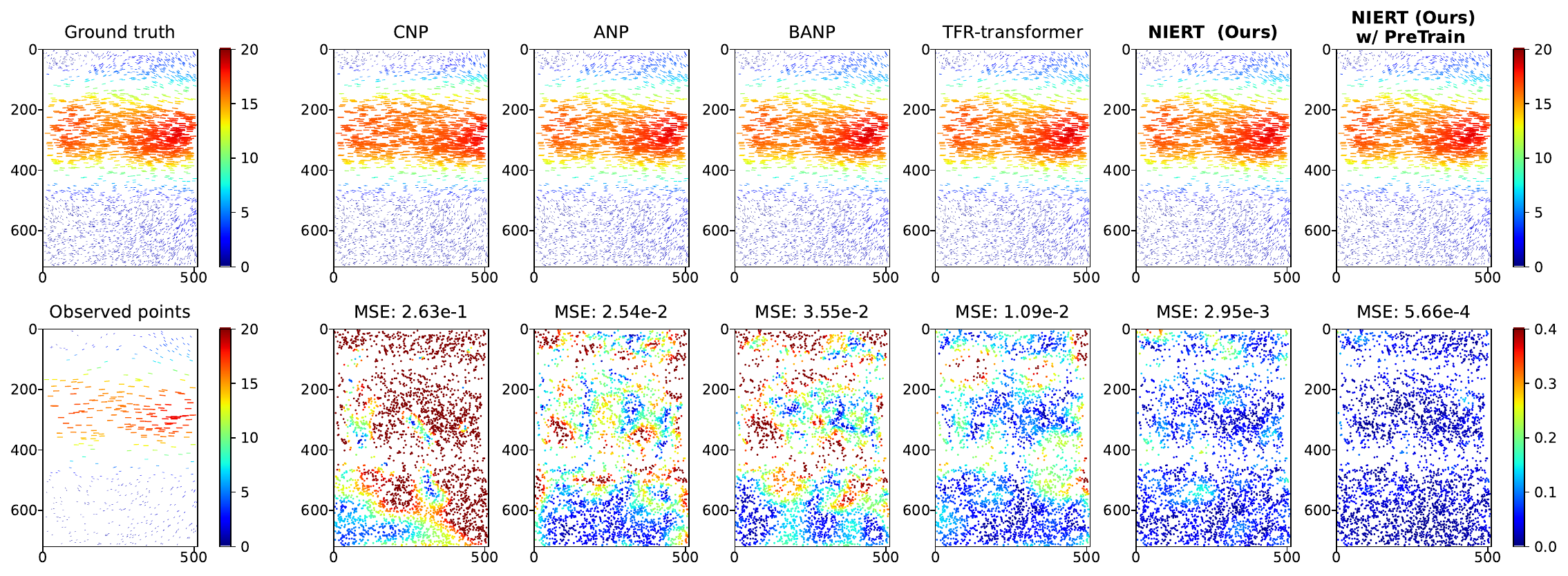} }} \\
    \caption{Additional cases from PTV test set}%
    \label{fig:more_ptv}%
\end{figure}

\newpage
\subsection{Contribution analysis of observed points for interpolation}
\label{subsec:contrib}

\begin{figure}[htbp!]
    \centering
    {{\includegraphics[width=15cm]{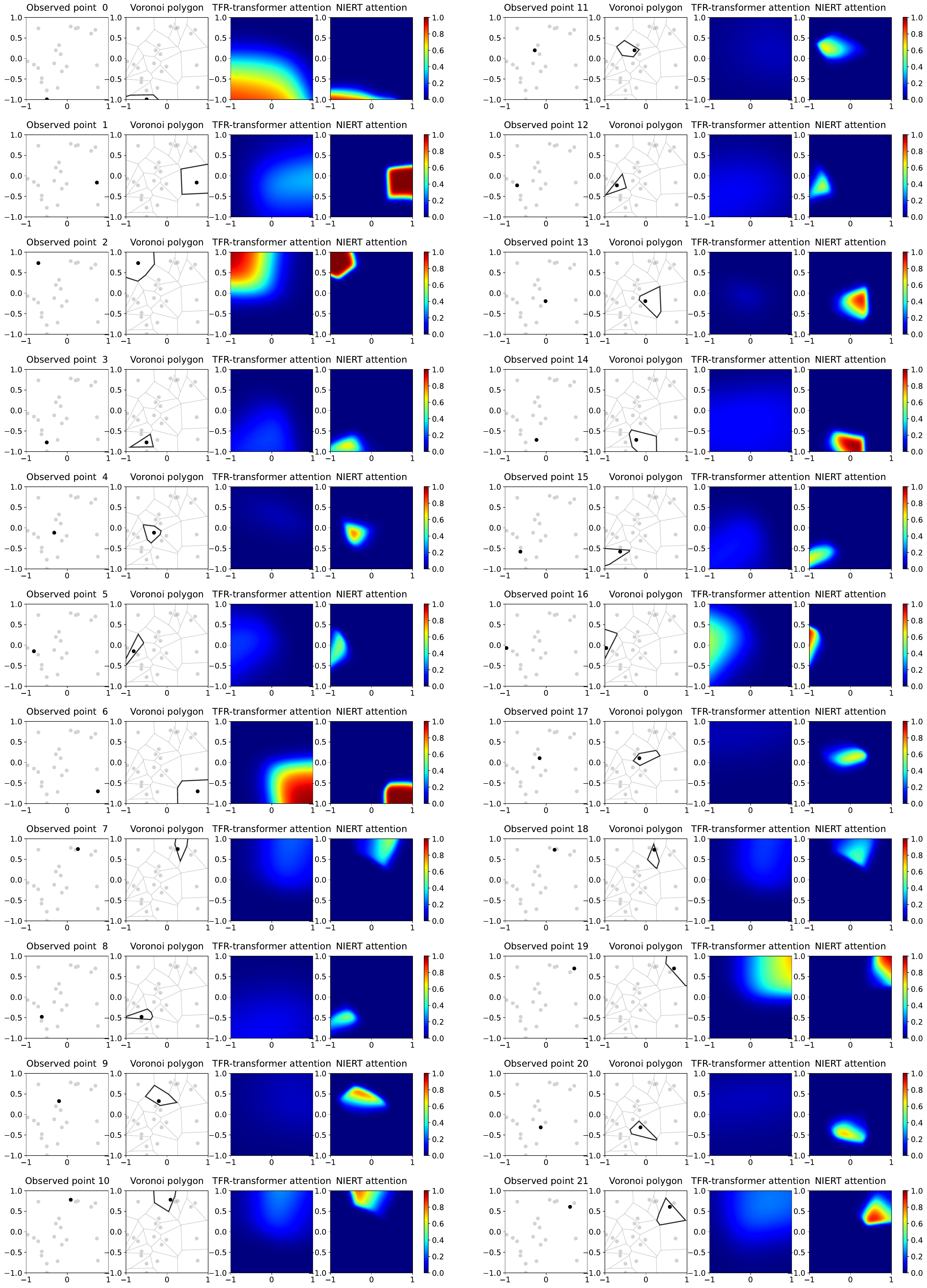} }}
    \caption{All observed points and the corresponding extracted attention response}%
    \label{fig:attention_all}%
\end{figure}

As supplements to Figure~6, all observed points' attention weights extracted from the final attention layer of NIERT and TFR-transformer are visualized in Figure~\ref{fig:attention_all}. 

The contributions by observed points are quite imbalanced using TFR-transformer. {In particular, the intensity and area of the contributions of the five observed points 0,2,6,16,19 cover almost the entire domain, while the remaining 17 observed points only produce negligible contribution.}
In contrast, when using NIERT, contributions by an observed point are much more local and thus targeted and all observed points have contributions to interpolation. This shows that NIERT can fully exploit the relationship between observed points and target points.

\end{document}